\documentclass[conference]{IEEEtran}
\IEEEoverridecommandlockouts

\usepackage{cite}
\usepackage{amsmath,amssymb,amsfonts}
\usepackage{algorithmic}
\usepackage{graphicx}
\usepackage{textcomp}
\usepackage{xcolor}
\usepackage{amssymb}
\usepackage{amsfonts}
\usepackage{colortbl}
\usepackage{enumerate}
\usepackage{multirow}
\usepackage{makecell}
\usepackage{threeparttable}
\usepackage{amsthm}
\usepackage{booktabs}
\usepackage{graphicx} 
\usepackage{algorithm}
\usepackage{algorithmic}
\usepackage{pifont}
\usepackage{xcolor}
\usepackage{hyperref}
\usepackage{array}
\usepackage{colortbl}
\def\BibTeX{{\rm B\kern-.05em{\sc i\kern-.025em b}\kern-.08em
    T\kern-.1667em\lower.7ex\hbox{E}\kern-.125emX}}
\begin{document}

\title{Federated Reasoning Distillation Framework with Model Learnability-Aware Data Allocation
}
\author{
Wei Guo$^{a,\dagger}$, Siyuan Lu$^{b,\dagger}$, Xiangdong Ran$^{c}$, Yiqi Tong$^{a}$, Yikun Ban$^{a}$, Zelong Xu$^{d}$, Jing Fan$^{a}$, \\ Zixuan Huang$^{a}$, Xiao Zhang$^{e,*}$, Zhaojun Hu$^{f}$, Fuzhen Zhuang$^{a,*}$ \\
$^{a}$\textit{Beihang University, Beijing 100191, China} \\
$^{b}$\textit{Heilongjiang University, Harbin 150080, China} \\
$^{c}$\textit{Beijing University of Civil Engineering and Architecture, Beijing 100044, China} \\
$^{d}$\textit{Yanshan University, Qinhuangdao 066104, China} \\
$^{e}$\textit{Shandong University, Qingdao 266237, China} \\
$^{f}$\textit{Renmin University of China, Beijing 100872, China} \\
\thanks{$\dagger$Equal contribution.}
\thanks{*Fuzhen Zhuang and Xiao Zhang are corresponding authors. Email: zhuangfuzhen@buaa.edu.cn, xiaozhang@sdu.edu.cn}
}
\maketitle

\begin{abstract}
Data allocation plays a critical role in federated large language model (LLM) and small language models (SLMs) reasoning collaboration. Nevertheless, existing data allocation methods fail to address an under-explored challenge in collaboration: \textbf{bidirectional model learnability gap}, where client-side SLMs cannot identify high-reward samples matching their learnability constraints for effective knowledge transfer from LLMs, while LLMs struggle to select samples contributing novel knowledge beyond their existing data. Furthermore, these collaboration frameworks face another key challenge: \textbf{domain-agnostic reasoning transfer}, where existing reasoning transfer methods fail to flexibly adapt to the local domain data, preventing SLMs from effectively acquiring step-by-step reasoning abilities within from general LLM. To address these challenges, we propose \textit{LaDa}, a federated reasoning distillation framework with model \textit{l}earnability-\textit{a}ware \textit{d}ata \textit{a}llocation. It introduces a \textit{model learnability-aware data filter} that adaptively allocates high-reward samples based on the learnability gap between each SLM and LLM pair, effectively facilitating bidirectional knowledge transfer. We further design a \textit{domain adaptive reasoning distillation} method that aligns joint probabilities of reasoning paths on filtered high-reward samples through contrastive distillation learning between SLM and LLM, enabling SLM to capture underlying reasoning patterns under local data distribution. \textit{LaDa} operates as a plug-in module for existing collaboration frameworks, adapting knowledge transfer based on model learnability gaps. We provide theoretical convergence guarantees with $O(1/\sqrt{T})$ rate for classic collaboration frameworks enhanced with our methods and demonstrate up to 13.8\% accuracy improvements over state-of-the-art baselines through extensive experiments across four LLM-SLM collaborative scenarios on two widely-used datasets. Our code is available at \hyperlink{https://github.com/GUoGUoWi/LaDa}{https://github.com/GUoGUoWi/LaDa}.

\end{abstract}


\section{Introduction}
\begin{figure}[t]
  \centering
  \includegraphics[scale=0.60]{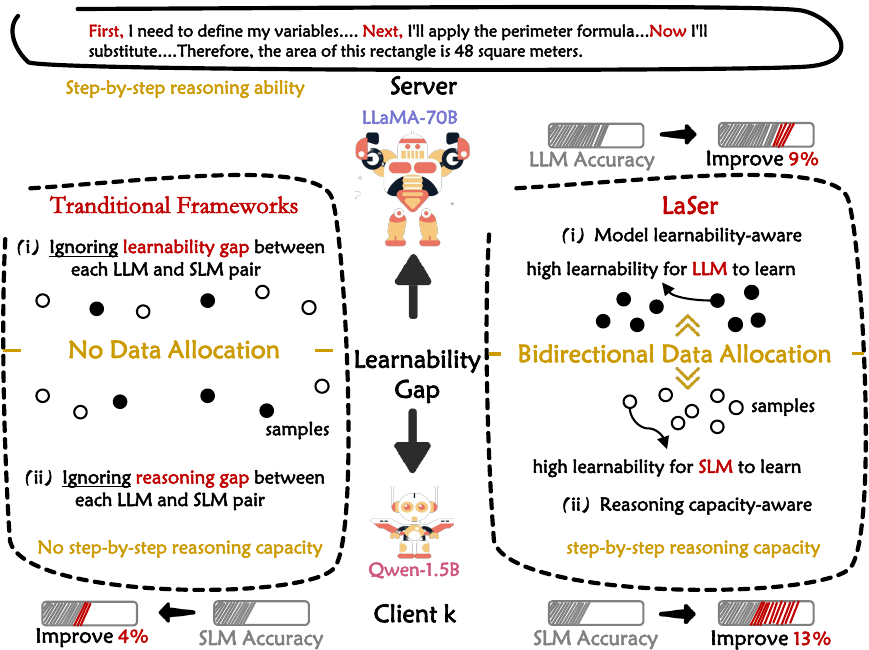}
  \caption{\textit{LaDa} identify an under-explored challenge in federated large-small model collaboration: bidirectional model learnability gap, where small langauge models (SLMs) and large language model (LLM) cannot identify their own high-reward samples matching their learnability constraint. Additionally, we address another key challenge: domain-agnostic reasoning transfer, where existing methods fail to flexible adapt to local domain data.} 
  \label{fig: intro}
\end{figure}
The collaborative paradigm between large language model (LLM) and small language models (SLMs) has emerged as a critical paradigm, enabling mutual enhancement through bidirectional knowledge transfer that addresses their complementary limitations \cite{wang2024comprehensive,fan2024fedcollm,chen2024data,zhu2025lslm,lu2024collaborative,zhang2025collm}. Data allocation \cite{chen2024data,fan2025fedmkt,lu2024collaborative,fan2024combining,wu2024agglomerative,zou2025contrastive,fan2025ppc,deng2023mutual,zoucontrastive,abiteboul2017research,10.1145/3626246.3654683,pedreira2024composable,song2025magnus,pei2023data} plays a pivotal role in this paradigm, critically governing the quality and effectiveness of knowledge transfer between LLM and SLMs with different scales and architectures.

Current research explores knowledge transfer between LLM and SLMs, covering both same-architecture methods \cite{fan2024fedcollm,hao2024hybrid,zhu2025lslm} and different-architecture methods \cite{fan2025fedmkt,zou2025contrastive,deng2023mutual,fan2025ppc,narayan2025cost,wu2024exploring,wu2024agglomerative,zhang2025collm,fan2024combining,zoucontrastive}. Same-architecture approaches include FedCoLLM \cite{fan2024fedcollm}, which employs lightweight adapters for knowledge transfer; EdgeFM \cite{hao2024hybrid}, which targets challenging tokens requiring LLM intervention; and CE-LSLM \cite{zhu2025lslm}, which enhances SLMs through LLM contextual guidance. However, these methods require identical model architectures and tokenizers. To overcome this limitation, different-architecture methods such as CTVLMs \cite{lu2024collaborative}, DS \cite{chen2024data}, FedMKT \cite{fan2025fedmkt}, FedAgg \cite{wu2024agglomerative}, CrossLM \cite{deng2023mutual}, PPC-GPT \cite{fan2025ppc}, and Cross-PLM \cite{zou2025contrastive} leverage knowledge distillation (KD) with public datasets to enable effective knowledge transfer between heterogeneous large-small models. MINIONS \cite{narayan2025cost} leverages code generation to decompose LLM tasks into parallel subtasks executed over public document data, with results aggregated by LLM.


Data allocation \cite{li2024llm} serves as the cornerstone for enhancing knowledge transfer in these large-small model collaborations. For example, FedMKT \cite{fan2025fedmkt} implements selective knowledge transfer by choosing logits with the smallest loss from multiple SLMs and applying them for distillation only when their loss is lower than LLM's own loss, ensuring only beneficial knowledge is transferred. WASP \cite{zoucontrastive} employs contrastive in-context learning with both high-quality and low-quality samples to guide LLMs in collaboratively generating high-quality data for training small task-specific SLM. DS \cite{chen2024data} filters distillation data using prediction confidence thresholds, mitigating impacts of distorted knowledge from large models.

Nevertheless, these data allocation methods overlook a critical yet under-explored challenge: \textbf{bidirectional model learnability gap}. On one hand, since SLMs display different and limited capacities to absorb knowledge from LLM \cite{li2025small}, client-side SLMs fail to identify high-reward samples matching their learnability constraints for effective knowledge transfer from LLM. On the other hand, LLM struggles to select samples that contribute novel knowledge beyond their existing learnable knowledge. For instance, a client-side 1.5B GEMMA cannot identify which distillation samples from a server-side 70B LLaDa match its learnability constraints, achieving better results with a similarly-sized 13B LLaDa. Conversely, a 70B LLaDa cannot distinguish which distillation samples from SLMs provide novel learnable knowledge, where a 1.5B GEMMA specializing in distinct domains may offer more valuable training samples than a 27B GEMMA with similar knowledge distributions. Notably, traditional data allocation methods \cite{li2022machine,zhang2024applications} designed for single models cannot be directly applied to large-small model collaborations, and fail to identify and mitigate their learnability gap challenges.

Moreover, we identify another critical challenge: \textbf{domain-agnostic reasoning transfer}, where existing reasoning transfer methods \cite{ranaldi2024self,qi2024mutual,fan2024combining,11127195,ranaldi2024aligning,ge2025knowtrans,nie2024knowledge,tong2025galaxyweaver,zhao2024chat2data} fail to flexible adapt to local domain data, preventing client-side SLMs from effectively acquiring step-by-step reasoning abilities according to local data distribution. Traditional approaches rely on supervised fine-tuning (SFT) with LLM-generated demonstrations \cite{shridhar2023distilling,ranaldi2024aligning,ranaldi2024self}, but suffer from limited generalization beyond provided examples. Recent works like rStar \cite{qi2024mutual}, which uses generator-discriminator mutual consistency for trajectory verification, and ZeroNL2SQL \cite{fan2024combining}, which combines SLM's structure identification with LLM's reasoning capabilities, are designed for general SLM and LLM and cannot dynamically adjust reasoning transfer when collaborating with different local data distribution in federated learning.

To address these challenges, we propose \textit{LaDa}, a federated reasoning distillation framework with model \textit{l}earnability-\textit{a}ware \textit{d}ata \textit{a}llocation. \textit{LaDa} includes: (i) \textit{model learnability-aware data filter}, that dynamically measures learnability gaps between each LLM-SLM pair by analyzing reasoning path differences across distillation samples, then adaptively allocates high-reward samples to optimize bidirectional knowledge transfer under various learnability gaps; and (ii) \textit{domain adaptive reasoning distillation}, that aligns joint probabilities of reasoning paths with contrastive distillation learning rather than overfitting to specific expression patterns, allowing each LLM and SLM pair to adaptively learn core reasoning patterns under their local data distribution.

The main contributions are summarized as follows:
\begin{enumerate}[\hspace{0em}(a)]
    \item We propose a plug-in for knowledge transfer in federated large-small model collaborations based on a public or synthetic dataset. To the best of our knowledge, we are the first to identify \textbf{bidirectional model learnability gap} in large-small model collaborations, a key yet under-explored challenge, where SLMs and LLM cannot identify high-reward samples matching their learnability gaps. Additionally, we dentify another key challenge: \textbf{domain-agnostic reasoning transfer}, where existing methods cannot flexible adapt to local domain data distribution. 

    
    \item We introduce \textit{LaDa}, a federated reasoning distillation framework with two key modules: (i) \textit{model learnability-aware data filter} that dynamically selects high-reward samples based on various different large-small model learnability gaps in federated learning; (ii) \textit{model adaptive reasoning distillation} that aligns joint probabilities of reasoning paths with contrastive distillation learning to transfer core reasoning capacity under local data distribution. We provide theoretical convergence guarantees with $O(1/\sqrt{T})$ rate for a classic collaboration framework enhanced with our proposed modules.
    \item Extensive experiments demonstrate that our proposed \textit{LaDa} achieves up to 13.8\% accuracy improvement over existing baselines across four diverse large and small model collaborative scenarios with varying model architectures and scales on two widely-used datasets.
    
\end{enumerate}

\section{Related Work}
\subsection{Federated Large-Small Model Collaboration}
Federated learning (FL) \cite{zhang2025federated,han2025securexgb,sun2024profit,jiang2024ofl,wei2024fedsm,zhu2024fedsq} enables a large language model (LLM) \cite{fang2025boosting,zhang2025delrec,sun2025gaussdb,luo2025natural,hussein2025large,li2150deepsearch,patel2025semantic,van2024dataloom,fan2024finding,jiang2025siriusbi,xiao2025cents,lao2025gptuner,fathollahzadeh2025catdb,santos2025interactive,giannakouris2024dbg,amer2023large} to train on distributed private data, addressing public data exhaustion and domain-specific limitations \cite{fan2025fedmkt,xu2025sequoia,fan2024fedcollm,deng2023mutual,bao2025towards,zhang2025online}. Due to resource constraints \cite{gong2022fedadmm,jiang2024clients,jiang2022fedmp,heinrich2024costream,aksoy2005information,mozafari2013performance,liu2024adaedge} and task heterogeneity \cite{yao2025joint,pan2023lumos,miao2021heterogeneity,yi2025pfedafm}, each client typically employs different small language models (SLMs), which drives extensive research \cite{fan2024fedcollm,fan2025fedmkt,zou2025contrastive,fan2025ppc,deng2023mutual,jajoo2025federated,fan2024pdss} towards federated large-small collaboration frameworks. Some works \cite{fan2024pdss,fan2025ppc} focus on unidirectional knowledge transfer between client-side SLMs and server-side LLM. For example, PDSS \cite{fan2024pdss} leverages LLMs to generate perturbed rationales for modified prompts to guide local SLM learning. Recent works enable bidirectional knowledge transfer. For instance, FedMKT \cite{fan2025fedmkt} introduces mutual knowledge distillation with public datasets, FedCoLLM \cite{fan2024fedcollm} adds secure aggregation for privacy, and CrossLM \cite{deng2023mutual} improves LLM generation quality through client feedback on synthetic data. However, these approaches overlook that larger models are not always optimal for knowledge transfer due to SLMs' capacity constraints. The proposed \textit{LaDa} addresses this through a learnability-aware data filter that enables adaptive bidirectional knowledge transfer tailored to specific LLM and SLM pairs.

\subsection{Data Allocation}
Data allocation \cite{li2022machine,li2024llm,albalaksurvey,fernandez2023large} acts as the critical mediator in knowledge transfer \cite{fan2024unicorn,fan2025fedmkt,chen2024data,hanmo2024effective,liu2025efficient,luo2025towards,yu2024unsupervised,li2024dual,he2024target,liu2020online,qiao2024feed,yang2021intellitag,zhang2023alt,yu2021windtunnel,trummer2023can,wu2023billion}. Nevertheless, traditional data allocation methods \cite{wang2020improved,liu2020self} cannot be directly applied across different LLM training stages \cite{albalaksurvey}. Consequently, current LLM-oriented data allocation methods target specific stages including pre-training \cite{yuan2023selecting,liang2025training}, instruction fine-tuning \cite{zhang2025survey,liu2024coachlm}, multi-task learning \cite{ge2025knowtrans,yu2025bigcity}, preference fine-tuning \cite{liu2025wepo,huang2025adaptive}, in-context learning \cite{fan2024cost,liang2025tailoring}, task-specific fine-tuning \cite{ge2025knowtrans,guo2024sample}, and domain adaptation \cite{zheng2024adapting,aycock2024topic}. These works are designed only for single LLMs and fail to improve knowledge transfer performance between SLM and LLM. To address this limitation, FedMKT \cite{fan2025fedmkt} filtered logits with the smallest loss from multiple SLMs and applying them for selective distillation only when their loss is lower than LLM's own loss. WASP \cite{zoucontrastive} employs contrastive in-context learning with both high-quality and low-quality samples to guide LLMs in collaboratively generating high-quality data for training small task-specific SLM. DS \cite{chen2024data} selected distillation data using prediction confidence thresholds, mitigating the impact of distorted knowledge from large models. Despite achieving promising results, these methods overlook the learnability gap and reasoning transfer challenge between large and small model collaborations.

\section{Problem Formulation}
We consider a federated large-small model collaboration scenario with one server hosting a large language model (LLM) $f_{\psi}$ parameterized by $\psi$, and $K$ clients each hosting a small language model (SLM) $g_{\phi_k}$ parameterized by $\phi_k$. Each client owns a local private dataset $\mathcal{D}_k$. Our method supports serving as a plug-in for existing federated large-small model collaboration frameworks under the mainstream setting where all clients share access to a public or synthetic dataset \cite{fan2025fedmkt,zou2025contrastive,lai2025tmlkd,wu2024blockchain,wu2023blocker,lu2024collaborative,chen2024data,wu2024agglomerative,deng2023mutual,fan2025ppc}. We denote this dataset as the distillation dataset $\mathcal{D}_p = \{x_{1},x_{2},...,x_{n}\}$.

\subsection{Problem Verification}
\begin{figure}[t] 
\centering
\includegraphics[width=1.0\linewidth]{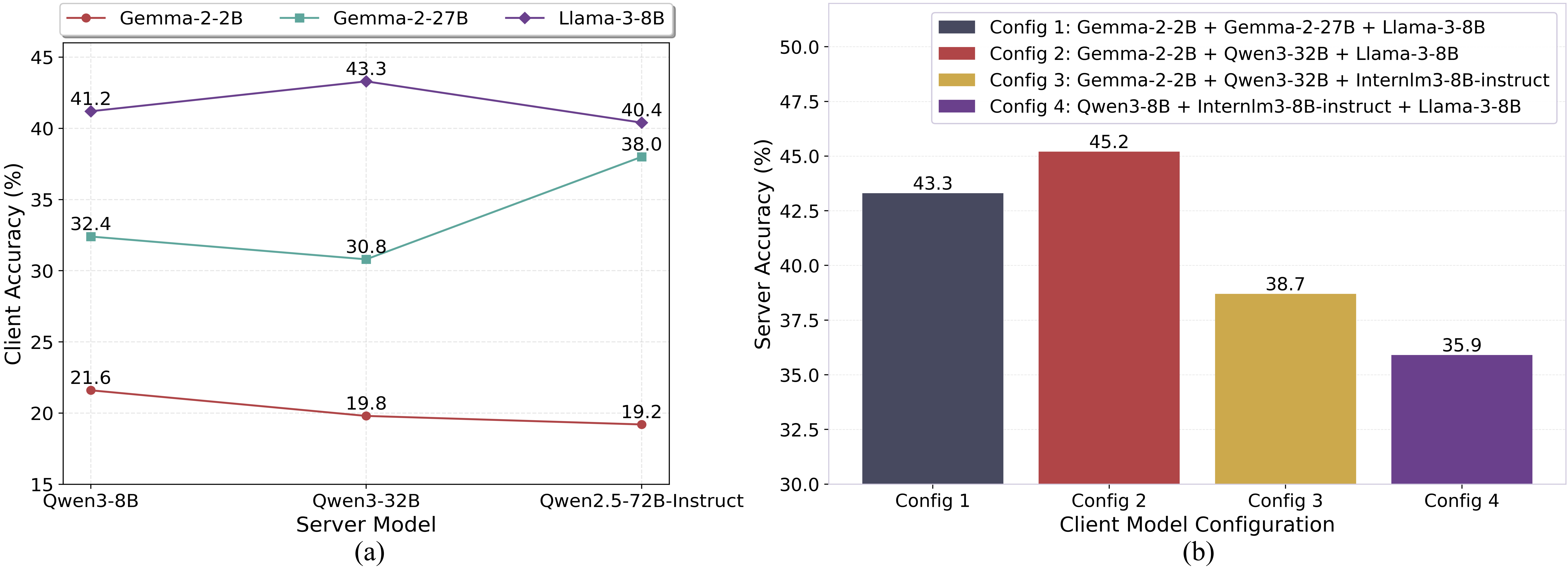}
\caption{Illustration of bidirectional model learnability gap: (a) client SLMs exhibit different learning capacities when collaborating with server LLMs of varying sizes; (b) Server LLM also shows varying knowledge absorption from different client SLM configurations.}
\label{fig:problem}
\end{figure}

To empirically validate bidirectional model learnability gaps, we conduct verification experiments on a traditional large-small model collaboration setting, employing classic knowledge distillation \cite{hinton2015distilling} on a public dataset to facilitate knowledge transfer. We perform validation on the full distillation dataset from two transfer directions: (i) fixing client-side SLMs while varying server-side LLM scales to evaluate SLM performance changes; (ii) fixing server-side LLM while varying client-side SLM configurations to examine LLM performance changes. The server-side LLM has three configurations: Qwen3-8B\footnote{\hyperlink{https://huggingface.co/Qwen/Qwen3-8B}{https://huggingface.co/Qwen/Qwen3-8B}}, Qwen3-32B\footnote{\hyperlink{https://huggingface.co/Qwen/Qwen3-32B}{https://huggingface.co/Qwen/Qwen3-32B}}, and Qwen2.5-72B-Instruct\footnote{\hyperlink{https://huggingface.co/Qwen/Qwen2.5-72B-Instruct}{https://huggingface.co/Qwen/Qwen2.5-72B-Instruct}}. The client-side SLMs include Gemma-2-2B\footnote{\hyperlink{https://huggingface.co/google/gemma-2-2b}{https://huggingface.co/google/gemma-2-2b}}, Gemma-2-27B\footnote{\hyperlink{https://huggingface.co/google/gemma-2-27b}{https://huggingface.co/google/gemma-2-27b}}, LLaDa-3-8B\footnote{\hyperlink{https://huggingface.co/meta-lLaDa/Meta-LLaDa-3-8B}{https://huggingface.co/meta-lLaDa/Meta-LLaDa-3-8B}}, Qwen3-32B, and Internlm3-8B-Instruct\footnote{\hyperlink{https://huggingface.co/internlm/internlm3-8b-instruct}{https://huggingface.co/internlm/internlm3-8b-instruct}}. All experiments are conducted on the MATHInstruct \cite{yue2023mammoth} dataset. Detailed experimental settings are provided in Sec. \ref{Experimental Setup}. As illustrated in Fig. \ref{fig:problem}(a), different client-side SLMs demonstrate non-monotonic performance patterns when collaborating with increasingly larger server-side LLMs, indicating that larger LLMs do not necessarily yield better client-side SLM performance when using the complete distillation dataset. Fig. \ref{fig:problem}(b) further reveals that LLM exhibits differential knowledge absorption across various client-side SLM configurations.

\subsection{Optimization Objective}
\textit{LaDa} aims to improve both server-side LLM and client-side SLMs' performance on the target task through bidirectional knowledge transfer, guided by a model learnability-aware data filter. Our bi-directional optimization objective includes:

\noindent\textbf{Server-side LLM:} LLM $f_{\psi}$ is optimized by learning diverse reasoning capacity and domain-specific knowledge from SLMs $f_{\theta_k}$ using distillation dataset $\mathcal{D}_p^s$ filtered by server. 
\begin{equation}\label{eq:server_loss}
    \min_{\psi} \mathcal{L}_{\text{server}}(\psi;\mathcal{D}_p^s,\{\phi_k\}_{k=1}^K).
\end{equation}

\noindent\textbf{Client-side SLMs:} Optimizing each SLM $f_{\theta_k}$ involves learning reasoning capacity from LLM $f_{\psi}$ through distillation dataset $\mathcal{D}_p^k$ filtered by each client $k$.
\begin{equation}\label{eq:client_loss}
     \min_{\phi_k} \mathcal{L}_{\text{client}}(\phi_k;\mathcal{D}_p^k,\psi).
\end{equation}

\begin{figure*}[t]
  \centering
  \includegraphics[scale=0.62]{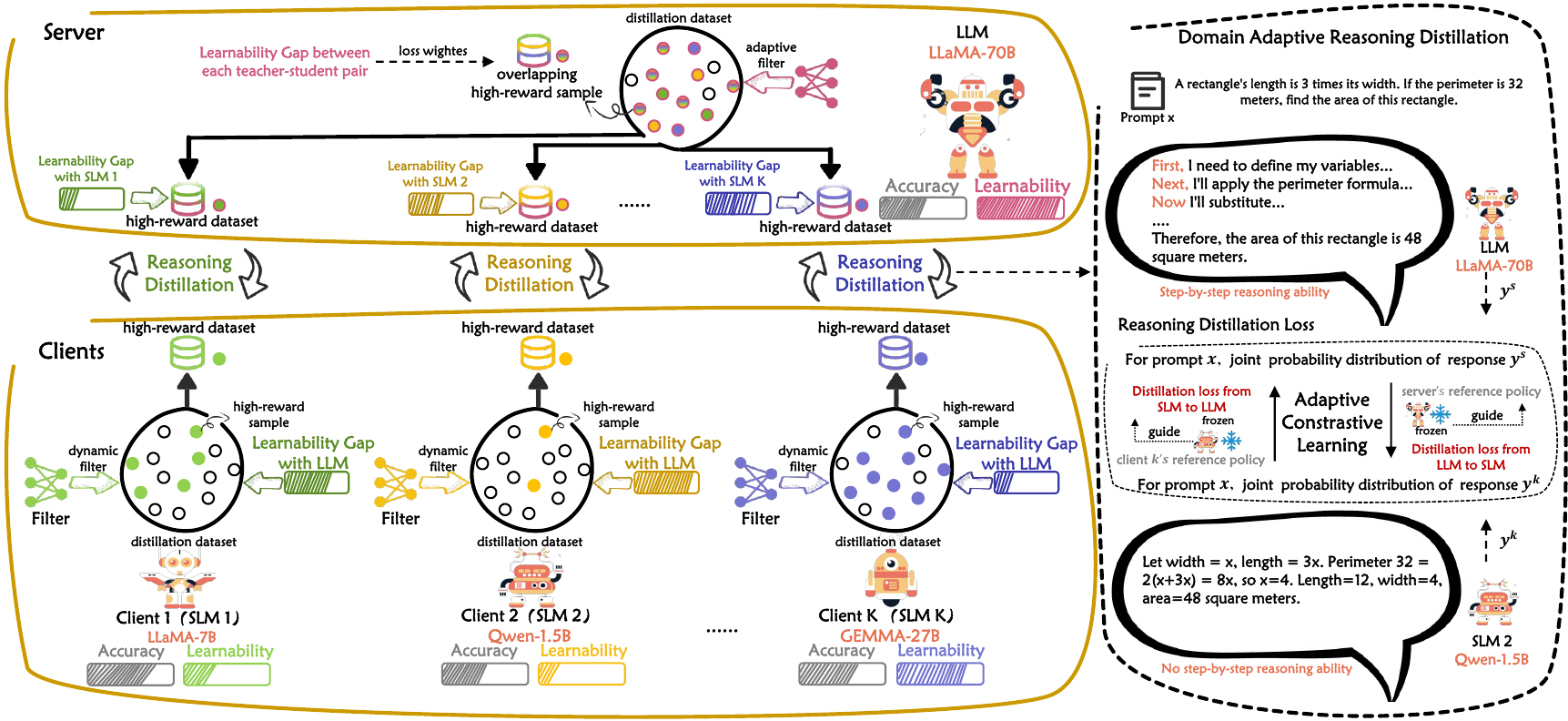}
  \caption{The overview of \textit{LaDa} framework. \textit{Model learnability-aware data filter}, which dynamically allocates high-reward samples based on bidirectional learnability gaps of each SLM and LLM pair; and \textit{domain adaptive reasoning distillation}, which aligns joint probabilities of reasoning paths through contrastive distillation learning, enabling domain adaptive reasoning capacity learning between each SLM and LLM pair.}
  \label{fig: framework}
\end{figure*}
\section{Methodology}
In this section, we introduce \textit{LaDa}, a federated reasoning distillation framework with model \textit{l}earnability-\textit{a}ware \textit{d}ata \textit{a}llocation, which addresses two critical challenges: \textbf{bidirectional model learnability gap} and \textbf{domain-agnostic reasoning transfer}. \textit{LaDa} comprises two key modules: \textit{model learnability-aware data filter} and \textit{domain adaptive reasoning distillation}. Notably, our modules functions as a plug-in for knowledge transfer in federated large-small model collaborations utilizing public or synthetic datasets \cite{lu2024collaborative,chen2024data,fan2025fedmkt,wu2024agglomerative,deng2023mutual,fan2025ppc,zou2025contrastive,zoucontrastive}. Since synthetic dataset methods primarily leverage synthetic data generated by server to match client data distributions, which remain accessible to both server and clients. For simplicity, we use public dataset $\mathcal{D}_p$ as distillation dataset to represent both settings throughout our methodology description. The overall architecture is illustrated in Fig. \ref{fig: framework}, with the complete training process detailed in Algorithm \ref{LaDa alg}.



\begin{algorithm}
\caption{Proposed Algorithm: \textit{LaDa}}
\begin{algorithmic}[t]\label{alg:ReDA}
\label{LaDa alg}
\REQUIRE Communication rounds $t$, batch $b$, client number $K$, and batch size $n_{t,b}$. 
\FOR{$t = 1, \ldots, T$}
    \FOR{$k = 1,2, \ldots, K$}
        \item[] \hspace{-0.5cm} $\triangledown$  \colorbox{lightgray}{\textbf{Obtain $\phi_{\text{ref}_k}$ and $\mathcal{D}_k^{t,b}$}}
        \IF{$t$ = 1}        
            \STATE Initialize $\phi_k$ with $\mathcal{D}_k$ to obtain $\phi_{\text{ref}_k}$ \ding{202}
            \STATE Initial high-reward samples $\mathcal{D}_k^{t}$ \ding{203}        
            \ENDIF
        \STATE Obtain batch data $\mathcal{D}_k^{t,b}$
        \item[] \hspace{-0.5cm} $\triangledown$ \colorbox{lightgray}{\textbf{Distillation Forward with} $\mathcal{D}_k^{t,b}$}\\
        \texttt{\# Standard Forward Pass}
        \STATE Compute Loss $\mathcal{L}^{\text{mard}}_k(\mathcal{D}_k^{t,b}; \phi_k^{t,b})$\\
        \item[] \hspace{-0.5cm} $\triangledown$  \colorbox{lightgray}{\textbf{Filter Training}}\\
        \texttt{\# Compute Sample Reward}
        \STATE Compute $r^{\text{round}}$ via Eq.\ref{eq:reward round} \texttt{\# Round} \ding{204}
        \STATE Compute $r^{\text{batch}}$ via Eq.\ref{eq:reward batch} \texttt{\# Batch} \ding{205}
        \STATE Compute $r^{\text{total}}$ via Eq.\ref{eq:reward final} \texttt{\# Total} \ding{206}\\
        \texttt{\# Update ExploitNet of Filter}
        \STATE Compute $\mathcal{L}^p$ and update $\theta^p_{t,b}$ via Eq.\ref{eq:update exploitor} \ding{207}\\
        \texttt{\# Update ExploreNet of Filter}
        \STATE Compute $\mathcal{L}^{q}$ and update $\theta^q_{t,b}$ via Eq.\ref{eq:update explorator} \ding{208}
        \item[] \hspace{-0.5cm} $\triangledown$  \colorbox{lightgray}{\textbf{Sample Filtering}}\\
        \texttt{\# Estimate Reward for Each Sample}
        \FOR{$i=1,2,\ldots,n_{t,b}$}
            \STATE Predict sample reward $\hat{r}_{t,b+1,i}$ via Eq.\ref{eq:estimate reward} \ding{209}
            \ENDFOR \\
        \texttt{\# Choose $\mathcal{D}_k^{t,b+1}$}
        \STATE $\mathcal{D}_k^{t,b+1} = \text{Top-}K_{i \in n_{t,b+1}} \hat{r}_{t,b+1,i}$ 
        \item[] \hspace{-0.5cm} $\triangledown$ \colorbox{lightgray}{\textbf{Distillation Backward with} $\mathcal{D}_k^{t,b+1}$}\\
        \texttt{\# Update local SLM parameters $\phi_k^{t,b+1}$}
        \STATE $\phi_k^{t,b+1} = \phi^{t,b}_k - \eta \nabla_{\phi^{t,b+1}_k} \mathcal{L}^{\text{mard}}_{k}(\mathcal{D}_k^{t,b}; \phi_k^{t,b+1})$ \ding{210}
    \ENDFOR
    \item[] \hspace{-0.5cm} $\triangledown$  \colorbox{lightgray}{\textbf{Server's LLM Updating}}\\
    \STATE Obtain filtered high-reward samples $D_s^{t,b}$ of server
    \STATE $\psi^{t,b}_s = \psi^{t,b+1}_s - \eta \nabla_{\psi^{t,b+1}_s} \mathcal{L}^{\text{mard}}_s(\{\mathcal{D}_s^{t,b}\}_{k=1}^K; \psi^{t,b+1}_s)$
\ENDFOR

\end{algorithmic}
\end{algorithm}

\subsection{Domain Adaptive Reasoning Distillation}
\label{sec:reasoning distillation loss}
Different from existing knowledge distillation (KD) methods \cite{zhang2020reliable,liu2024lighttr,yao2023resuformer,wang2023incremental,tang2024substructure,yu2021windtunnel}, \textit{LaDa} introduces a novel distillation method for reasoning transfer to facilitates the domain adaptive transfer of reasoning capabilities from server-side LLM to client-side SLMs. Inspired by direct preference optimization (DPO) \cite{rafailov2023direct}, our method utilizes reasoning responses $y^s$ and $y^k$ from LLM and SLM to the same problem $x$ as preferred and non-preferred answers in DPO. We then design a novel distillation loss $\mathcal{L}^{\text{mard}}$ that enables each SLM to adaptively align its reasoning path distribution with that of the LLM through contrastive learning.

Specifically, given a sample $a = (x,y^k,y^s)$ in distillation dataset $\mathcal{D}_p$, $x$ represents input prompt and $y^k$, $y^s$ denote reasoning answers to $x$ generated by client $k$ and server $S$, respectively. According to reward definition based on probability distribution in \cite{rafailov2023direct}, it is incorporated into the Bradley-Terry ranking objective \cite{bradley1952rank}, and we obtain probability $p(y^s>y^k|x)=\sigma(r(x,y^s)-r(x,y^k))$, where $\sigma$ denotes the logistic function. Consequently, we define client-side loss function as:
\begin{equation}
    \scriptsize
    \mathcal{L}^{\text{mard}}_{\text{client}}(a;\phi_k) = -\mathbb{E}_{a\sim \mathcal{D}^k_p}\Big[\text{log}\sigma\Big(\beta\big(\text{log}\frac{\phi_{k}(y^s|x)}{\phi_{\text{ref}_k}(y^s|x)}-\text{log}\frac{\phi_{k}(y^k|x)}{\phi_{\text{ref}_k}(y^k|x)}\big)\Big) \Big],
\label{eq:client update}
\end{equation}
where $\phi_{\text{ref}_k}$ is the supervised fine-tuned reference policy on local dataset (\ding{202}). $\mathcal{D}^k_p$ contains high-reward samples selected from $\mathcal{D}_p$ by client $k$, with its size reflecting learnability gap between client $k$'s SLM and server-side LLM, as detailed in Sec.\ref{sec:scheduler}. $\beta$ controls the regularization strength, balancing model divergence from the reference policy when learning from reasoning responses. Based on this, each client fine-tunes its local model using reasoning response set generated by server's LLM on dataset $\mathcal{D}_p^k$. The fine-tuning process focuses on fitting joint probability distributions of output tokens between each SLM and LLM, guiding client model $\phi_k$ to approximate reasoning patterns exhibited in LLM.

Similarly, given architectural and domain knowledge differences between LLM and each SLM, LLM can learn new reasoning patterns from these diverse SLMs. When using a synthetic dataset as distillation dataset $\mathcal{D}_p$ that better match local data distributions, LLM can also acquire more domain-specific knowledge from private data. This allows LLM to expand beyond its original training knowledge. To distill knowledge from SLMs to LLM, the server also selects its own high-reward samples with each SLM from $\mathcal{D}_p$ (Sec.\ref{sec:scheduler}). Unlike the client-side selection where each client works with a single high-reward sample set, the server receives multiple high-reward sample sets from $K$ clients when learning from their SLMs, which may contain overlapping samples across different clients. For these overlapping samples $\mathcal{D}_p^{s,o}$, we apply a weighted loss summation during LLM fine-tuning, while for non-overlapping samples $\mathcal{D}_p^{s,no}$, we directly use the loss $\mathcal{L}^{\text{mard}}$ computed for each LLM-SLM pair:
\begin{equation}
\tiny
    \mathcal{L}^{\text{mard}}_{\text{server}}(a;\psi) =\left\{
    \begin{aligned}
    & -\mathbb{E}_{a\sim \mathcal{D}_p^{s,o}}\Big[\sum^{K}_{k=1}w_k\text{log}\sigma\Big(\beta\big(\text{log}\frac{\psi(y^k|x)}{\psi_{\text{ref}}(y^k|x)}-\text{log}\frac{\psi(y^s|x)}{\psi_{\text{ref}}(y^s|x)}\big)\Big) \Big],\\
    &-\mathbb{E}_{a\sim \mathcal{D}_p^{s,no}}\Big[\text{log}\sigma\Big(\beta\big(\text{log}\frac{\psi(y^k|x)}{\psi_{\text{ref}}(y^k|x)}-\text{log}\frac{\psi(y^s|x)}{\psi_{\text{ref}}(y^s|x)}\big)\Big) \Big],
    \end{aligned}
    \right. 
\label{eq:server update}
\end{equation}
where $\psi_{\text{ref}}$ represents the LLM's original pre-trained parameters, which remain fixed along with the SLM reference policy $\phi_{\text{ref}_k}$. These references prevent the optimized models from diverging excessively from their original distributions.

\subsection{Model Learnability-Aware Data Filter}
\label{sec:scheduler}
To address model learnability gaps between the server-side LLM and each client-side SLM, we introduce a sample filter that selects high-reward samples tailored to learning capacity of each LLM and SLM pair, preventing excessive knowledge transfer in collaboration process. While existing works \cite{li2025small,fan2025fedmkt,zou2025contrastive} control knowledge transfer by filtering low-quality samples, they fail to adaptively select high-reward samples across diverse LLM-SLM pairs with varying learnability gaps. To overcome this limitation, we propose a novel model learnability-aware data filter based on the exploration-exploitation paradigm \cite{vcrepinvsek2013exploration,ban2024neural,chai2022selective,gubner2022excalibur,khayati2024imputevis}. Since learnability-aware filtering decision for each sample is a sequential process where both LLM and SLM continuously evolve, and their internal states and learnability dynamically change during training, introducing uncertainty in data filtering. Naturally, we address this challenge through the exploration-exploitation framework.

Specifically, at each round $t$, each client filter $U_k$ and server filter $U_s$ adaptively choose high-reward samples $\mathcal{D}_p^{k,t,b}$ and $\mathcal{D}_p^{s,t,b}$ from $\mathcal{D}_p$ that match current model learnability gap after each batch $t$ update. Notably, as high-reward sample filtering occurs at both server and client with similar methods, we use client-side implementation for illustration.

\textbf{Reward definition.} To quantify each sample's reward value, we design a reward function $r$ that measures performance gains through model loss variation \cite{lecun1989optimal,di2025survey,chen2025compressgnn} at each update, leveraging insights from the learning trajectory. This reward serves as the supervisory signal to guide filter $U_k$ training through two components: round-level and batch-level.

For round-level reward, it measures reasoning distillation loss reduction before and after training with selected samples. We use round-average loss to approximate the expected loss across the entire data distribution. Formally, at communication round $t$, the model parameters are updated from $\phi^{t-1}_k$ to $\phi^{t}_k$ by training on the previously selected subset $\mathcal{D}_p^{k,t-1}$ containing $n_{t-1}$ samples. Using the updated model $\phi^{t}_k$, the filter then selects a new subset $\mathcal{D}_p^{k,t}$ with $n_{t}$ samples for the current round. The round-level reward is defined as (\ding{204}):
\begin{equation}
\begin{aligned}
    r^{\text{round}}=
    \frac{\overbrace{\sum\limits^{n_{t-1}}_{i=1}e^{\mathcal{L}^{\text{mard}}_{\text{client}}(a_{t-1,i};\phi^{t-1}_k)}}^{M}-\overbrace{\sum\limits^{n_{t}}_{j=1}e^{\mathcal{L}^{\text{mard}}_{\text{client}}(a_{t,j};\phi^{t}_k)}}^{N}}{\text{max}(\sum\limits^{n_{t-1}}_{i=1}e^{\mathcal{L}^{\text{mard}}_{\text{client}}(a_{t-1,i};\phi_{k}^{t-1})},\sum\limits^{n_t}_{i=1}e^{\mathcal{L}^{\text{mard}}_{\text{client}}(a_{t,i};\phi_{k}^{t})})},
\end{aligned}
\label{eq:reward round}
\end{equation}
where term $M$ evaluates $\phi_{k}^{t-1}$'s performance on $\mathcal{D}_p^{t-1}$, term $N$ measures $\phi_{k}^{t}$'s performance on the newly selected set $\mathcal{D}_k^{t}$. We use exponentiated loss values with normalization to increase reward sensitivity. $r^{\text{round}}$ represents model improvements following filter decisions at round $t$ for each example.

Similar to round-level reward, at batch $b$ with $n_{t,b}$ samples of communication round $t$, batch-level reward is define as (\ding{205}):
\begin{equation}
\begin{aligned}
    r^{\text{batch}}=
    \frac{\sum\limits^{n_{t,b}}_{i=1}e^{\mathcal{L}^{\text{mard}}_{\text{client}}(a_{t,b-1,i};\phi_{k}^{t,b-1})}-\sum\limits^{n_{t,b}}_{i=1}e^{\mathcal{L}^{\text{mard}}_{\text{client}}(a_{t,b,i};\phi_{k}^{t,b})}}{\text{max}(\sum\limits^{n_{t,b}}_{i=1}e^{\mathcal{L}^{\text{mard}}_{\text{client}}(a_{t,b-1,i};\phi_{k}^{t,b-1})},\sum\limits^{n_{t,b}}_{i=1}e^{\mathcal{L}^{\text{mard}}_{\text{client}}(a_{t,b,i};\phi_{k}^{t,b})})}.
\end{aligned}
\label{eq:reward batch}
\end{equation}

Finally, we integrate round-level and batch-level rewards to compute the final reward for each sample $a_{t,b,i}$ (\ding{206}):
\begin{equation}
\begin{aligned}
    r^{\text{total}}
    =\alpha \sigma (r^{\text{round}})+(1-\alpha) \sigma (r^{\text{batch}}),
\end{aligned}
\label{eq:reward final}
\end{equation}
where $\alpha\in[0,1]$ are hyperparameters balancing these reward components, and $\sigma(\cdot)$ is the sigmoid function.

\textbf{Adaptive Sample Filter.}
Sample reward indicates knowledge transfer preference between specific LLM and SLM pairs. We use it as supervisory signals to train a flexible filter that predicts sample reward across different model pairs, addressing model learnability gaps for effective knowledge transfer. As sample filter involves sequential decisions with dynamically updating models, our approach naturally incorporates the exploration-exploitation dilemma.

Specifically, the filter $U_k$ combines an encoder with two specialized networks: ExploitNet ($U^p_k$) and ExploreNet ($U^q_k$). ExploitNet predicts sample rewards by mapping inputs to observed rewards, while ExploreNet estimates prediction uncertainty and adds an exploration bonus. This design balances exploitation and exploration during sample selection, following principles from UCB \cite{qin2014contextual,wen2015efficient} and Thompson Sampling algorithms \cite{zhang2020neural,zhao2023comparative,vaidya2021leveraging}. Given the $b$-th batch input $x_{t,b,i}$ in round $t$, the exploitation network $U^p_k$ is implemented as a fully connected feedforward neural network with residual connections, denoted by $U^p_k(x_{t,b,i};\theta^p_{t,b})$. After receiving the observed reward $r^{\text{total}}$, the parameters $\theta^p_{t,b}$ are updated by (\ding{207}):
\begin{equation}
\small
\begin{aligned}
    \mathcal{L}^p(\mathcal{D}_k^{t,b},\theta^p_{t,b})&=\frac{1}{2n_{t,b}}\sum\limits^{n_{t,b}}_{i=1}\big[U^p(x_{t,b,i};\theta^p_{t,b})-r^{\text{total}}\big]^2.
\end{aligned}
\label{eq:update exploitor}
\end{equation}
Next, in each batch $b$, we construct the input to $U^q_k$ by concatenating intermediate hidden states of $U^p_k(x_{t,b,i};\theta^p_{t,b-1})$ along the last dimension, denoted by $h^{p}_{t,b,i}$. This design enables the exploration module to take into account the internal states of the ExploitNet when making exploration decisions. The ExploreNet $U^q_k$ is also a fully connected feedforward neural network with residual connections. After receiving the observed reward $r^{\text{total}}$ in batch $b + 1$, the label for training $U^q_k$ is the difference between observed reward and $U^p(\cdot;\theta^p_{t,b})$ for uncertainty estimation. The ExploreNet parameters $\theta^q_{t,b}$ are then updated by (\ding{208}):
\begin{equation}
\scriptsize
\begin{aligned}
    \mathcal{L}^q(\mathcal{D}_k^{t,b},\theta^q_{t,b})&=\frac{1}{2n_{t,b}}\sum\limits^{n_{t,b}}_{i=1}\big[U^q_k(h^{p}_{t,b,i};\theta^q_{t,b})-(r^{\text{total}}-U^p(x_{t,b,i};\theta^p_{t,b}))\big]^2.
\end{aligned}
\label{eq:update explorator}
\end{equation}
Finally, the overall reward estimation for each sample is (\ding{209}): 
\begin{equation}
    \hat{r}_{t,b,i}(x_{t,b,i};\theta^p_{t,b},\theta^q_{t,b})=U^p_k(x_{t,b,i};\theta^p_{t,b})+\lambda U^q_k(h^p_{t,b,i};\theta^q_{t,b}),
\label{eq:estimate reward}
\end{equation}
where $\lambda$ is a tunable hyperparameter controlling the strength of exploration. After training the filter with sample rewards as supervision signals, the updated filter selects samples with the predicted rewards $\hat{r}>0.5$ as the preferred samples for the current batch. These selected samples are then used to query reasoning responses from LLM and SLM, which will be utilized for the current round of target model training (\ding{210}). Notably, since the filter is not trained in the first round, we employ clustering to identify sample prototypes as high-reward samples for the initial selection (\ding{203}). 

\section{Convergence Analysis}\label{sec:conver}
In this section, we provide a convergence analysis of our proposed algorithm. Under the standard bounded variance assumption (cf. Assumption 4), the effect of the sample selector is fully captured through the stochastic gradient it yields. That is, selecting samples via a sample selector and directly sampling them randomly are equivalent from a convergence analysis perspective. Therefore, for simplicity, we do not explicitly incorporate the sample selection mechanism in our analysis. For the sake of clarity, we introduce some notations. We denote $\Psi:=\{\psi \}$ and $\Phi:=\{\phi_1,\phi_2,\cdots,\phi_{K-1},\phi_K \}$. We define $L_{{\rm client}}( {\Theta}, {\mathcal{E}};\zeta):=l_{{\rm client}}({\Theta}, {\mathcal{E}};\zeta) + \frac{\varphi}{2} \| \mathcal{E}\|_F^2$, where $\zeta$ denotes a single sample from $K$ clients and $l_{{\rm client}}$ is the base loss function for a single sample in the clients. The optimization problems in \eqref{eq:server_loss} and \eqref{eq:client_loss} thus form a bilevel optimization problem, which can be formulated as follows.

\begin{equation}
    \begin{aligned}
         \min_{{\Psi}} F({\Psi})&:= \mathcal{L}_{{\rm server}}( {\Psi}, {\Phi}^*( {\Psi}))=\mathbb{E}_{\xi}[L_{{\rm server}}( {\Psi}, {\Phi}^*( {\Psi});\xi)]\\
         &=\frac{1}{n}\sum_{i=1}^{n}L_{{\rm server}}( {\Psi}, {\Phi}^*( {\Psi});\xi_i),\\
     \text{s.t.} \quad \Phi^*(\Psi)&=\arg\min_{\Phi} \mathcal{L}_{{\rm client}}( \Psi,\Phi)=\mathbb{E}_{\zeta}[L_{{\rm client}}( {\Psi}, {\Phi};\zeta)] \\
     &= \frac{1}{n}\sum_{i=1}^{n}L_{{\rm client}}( {\Psi}, {\Phi};\zeta_i),
    \end{aligned}\label{bilevel_opt}
\end{equation}
where $L_{{\rm server}}$ and $L_{{\rm client}}$ represent the loss functions for a single sample in the server and clients, respectively. 

Let $\mathcal{Z}:= \{\Psi, \Phi\}$ denote all model parameters, we then make some standard assumptions \cite{ghadimi2018approximation,ji2020convergence,rajput2020closing} on $L_{{\rm server}}$ and $L_{{\rm client}}$ for our bi-level optimization problem \eqref{bilevel_opt}.

\noindent \textbf{Assumption 1. (Lipschitz Condition)} The loss function $L_{{\rm server}}(\mathcal{Z};\xi)$ is $L_1$-Lipschitz for any given $\xi$.

\noindent \textbf{Assumption 2. (Smoothness)} The loss functions $L_{{\rm server}}(\mathcal{Z};\xi)$, $L_{{\rm client}}\allowbreak (\mathcal{Z};\xi)$ and $l_{{\rm client}}(\mathcal{Z};\xi)$ are $L_2$-smooth, $L_2$-smooth and $L$-smooth for any given $\xi$ and $\zeta$, respectively.

\noindent \textbf{Assumption 3. (Lipschitz Condition for Second Derivatives)} The second derivatives $\nabla_{\Psi}\nabla_{\Phi}L_{{\rm client}}(\mathcal{Z};\zeta)$ and $\nabla_{\Phi}^2 L_{{\rm client}}(\mathcal{Z};\zeta)$ are $L_3$-Lipschitz and $L_4$-Lipschitz for any given $\zeta$, respectively.

\noindent \textbf{Assumption 4. (Bounded Variance)} The stochastic gradient $\nabla L_{{\rm client}}\allowbreak (\mathcal{Z};\zeta)$ has a bounded variance $\sigma^2$, i.e., for any $\mathcal{Z}$, we have:
\begin{equation*}
    \mathbb{E}_{\zeta} \|\nabla L_{{\rm client}}(\mathcal{Z};\zeta) - \nabla \mathcal{L}_{{\rm client}}(\mathcal{Z})\|^2 \leq \sigma^2.
\end{equation*}

\noindent \textbf{Assumption 5. (Bounded Domain)} The parameter $\Phi$ is in a bounded domain with a diameter $\Delta$, i.e., for any $\Phi_1$ and $\Phi_2$, we have:
\begin{equation*}
    \| \Phi_1 - \Phi_2\|\leq \Delta.
\end{equation*}



For simplicity, we assume that the batch sizes of the server’s batch data and the clients’ batch data are the same, and we denote this common batch size by $N_B$. 

\noindent \textbf{Proof Sketch}
To explore the essential insights, we first bound the difference between the local parameter and the local optimal parameter at the $j$-th epoch (i.e. tracking error $\|\Phi_{t}^{j-1} - \Phi^*(\Psi_t^0) \|$). Then, the upper bound of the difference between the approximate gradient and the exact gradient (i.e. $\Big\| \frac{\partial \mathcal{L}_{\rm client}(\Psi_t^j, \Phi_t^\tau)}{\partial \Psi_t^j} - \nabla F(\Psi_t^j) \Big\|$) is provided by a proof technology called virtual updates (\cite{yang2022fastslowmo}).

To give the convergence analysis of our algorithm, we need to give some useful lemmas first. With $L$-smoothness of $l_{\rm client}(\Psi,\Phi;\zeta)$, we could give the strongly convexity property of $L_{\rm client}(\Psi,\Phi;\zeta)$.

\noindent\textbf{Lemma 1.} Under Assumption 2, suppose $\alpha:=-L+\varphi >0$, $L_{\rm client}(\Psi,\Phi;\zeta)$ is $\alpha$-strongly convex w.r.t $\Phi$. 

\noindent\textbf{Lemma 2.} Under Assumption 1 and 2, the derivatives $\nabla L_{\rm server} (\mathcal{Z}; \xi)$, $\nabla_{\Psi}\nabla_{\Phi}L_{\rm client}(\mathcal{Z};\zeta)$ and $\nabla_{\Phi}^2 L_{\rm client}(\mathcal{Z};\zeta)$ have bounded variances, i,e., for any $\mathcal{Z}$, we have
\begin{equation}
    \mathbb{E}_{\xi} \|\nabla L_{\rm server} (\mathcal{Z};\xi) - \nabla \mathcal{L}_{\rm server} (\mathcal{Z})\|^2 \leq L_1^2,\label{cls_bound}
\end{equation}
\begin{equation}
    \mathbb{E}_{\zeta} \|\nabla_{\Psi}\nabla_{\Phi}L_{\rm client} (\mathcal{Z};\zeta) - \nabla_{\Psi}\nabla_{\Phi}\mathcal{L}_{\rm client} (\mathcal{Z})\|^2 \leq L_2^2,\label{lv_bound1}
\end{equation}
\begin{equation}
    \mathbb{E}_{\zeta} \|\nabla_{\Phi}^2 L_{\rm client} (\mathcal{Z};\zeta) - \nabla_{\Phi}^2\mathcal{L}_{\rm client} (\mathcal{Z})\|^2 \leq L_2^2.\label{lv_bound2}
\end{equation}

\noindent\textbf{Lemma 3. (Lemma 2.2 in \cite{ghadimi2018approximation})} Under Assumptions 1, 2 and 3, $F (\Psi)$ is $L_0$-smooth where $L_0$ is given by
\begin{equation}
\small
    L_0:= L_2 +\frac{2L_2^2 + L_1^2 L_3}{\alpha}+ \frac{L_1L_2L_3+L_1L_2L_4+L_2^3}{\alpha^2}+\frac{L_1L_2^2L_4}{\alpha^3}.
\end{equation}

Tracking error $\mathbb{E}\|\Phi_t^{j-1} - \Phi^*(\Psi_t^0)\|$ is an important component in our convergence analysis. To give an upper bound on the tracking error, we utilized Lemma 9 in \cite{ji2021bilevel}.

\noindent\textbf{Lemma 4. (Lemma 9 in \cite{ji2021bilevel})} Under Assumptions 1, 2 and 4, with stepsize $\eta$ to be $\frac{2}{L_2 + \alpha}$, we have
\begin{equation}
\scriptsize
    \mathbb{E}\|\Phi_t^{j-1} - \Phi^*(\Psi_t^0)\|^2 \leq \left(\frac{L_2 - \alpha}{L_2 +\alpha}\right)^{2(j-1)}\mathbb{E}\|\Phi_t^{0} - \Phi^*(\Psi_t^0)\|^2 + \frac{\sigma^2}{L_2\alpha S}. \label{tracking_error}
\end{equation}

With the above lemmas, we could give the estimation property of the $\frac{\partial L_{\rm server} (\Psi_t^0,\Phi_t^\tau)}{\partial \Psi_t^0}$ approximating $\nabla F(\Psi_t^0)$.

\noindent\textbf{Proposition 1.} Under Assumptions 1-5, suppose $\alpha<L_2$ and let stepsize $\eta$ be $\frac{2}{L_2 + \alpha}$, we have
\begin{equation}
\small
    \begin{aligned}
        &\mathbb{E}\|\frac{\partial L_{\rm server} (\Psi_t^0,\Phi_t^\tau;\mathcal{S}_0^\prime)}{\partial \Psi_t^0} - \nabla F(\Psi_t^0)\|\leq (L_2 + \frac{L_2^2}{\alpha})[\left(\frac{L_2 - \alpha}{L_2 +\alpha}\right)^{\tau} \\
        &\sqrt{\Delta}+ \frac{\sigma}{\sqrt{L_2\alpha S}}]+L_1[\frac{L_2 (1-\frac{2}{L_2+\alpha} \alpha)^{\tau}}{\alpha}+\frac{1}{\alpha\sqrt{S}}(\frac{L_2^2}{\alpha}+L_2)\\
        &+\frac{\sigma}{\alpha\sqrt{L_2\alpha S}}(\frac{L_2L_4}{\alpha}+L_3)+\frac{2}{L_2+\alpha} (\frac{L_2L_4}{\alpha}+L_3)\sqrt{\Delta}\\
        &\frac{(1-\frac{2}{L_2+\alpha}\alpha)^\tau}{1-\frac{2}{L_2+\alpha}\alpha - \frac{L_2-\alpha}{L_2+\alpha}}]+\frac{L_1}{\sqrt{S}}.
    \end{aligned}\label{prop1}
\end{equation}

Finally, we present our convergence theorem.

\noindent\textbf{Theorem 1.} Under Assumptions 1-5, given a small tolerance $\epsilon$, define $\alpha:=-L+\varphi$, choose step size $\eta$ to be $\frac{2}{L_2 + \alpha}$, $N_B=\mathcal{O}(\frac{1}{\sqrt{\epsilon}})$, $\tau=\mathcal{O}(\log\frac{1}{\epsilon})$, $\eta^\prime <\frac{1}{2L_0}$ and suppose $\alpha<L_2$, we have:
\begin{equation}
\small
    \begin{aligned}
        \frac{1}{\tau^\prime T}&\sum_{t=0}^{T-1}\sum_{j=0}^{\tau^\prime-1}\|\nabla F(\Psi_t^{j})\|^2 \leq \frac{F(\Psi_0^0) - \inf_{\Theta}F(\Psi)}{\tau^\prime T (\frac{\eta^{\prime}}{2} -L_0 \eta^{\prime 2}) }\\
        &+(\eta^\prime + 2\eta^{\prime 2}L_0)L_2^2\Delta^2 \frac{\tau^\prime -1}{\tau^\prime}+\mathcal{O}(\epsilon),
    \end{aligned}\label{theorem_1}
\end{equation}
where $\mathcal{O}(\cdot)$ indicates that the function's growth rate is proportional to or slower than $\epsilon$. If we set $\tau^\prime=1$, convergence rate becomes $\mathcal{O}(\frac{1}{\sqrt{T}})$.

\begin{table*}[t]
  \centering
  \caption{Model configurations across four experimental scenarios.}
    \label{tab:1}
    \begin{tabular}{ccccccc}
      \toprule
      \multirow{2}{*}{Scenario} & \multicolumn{3}{c}{SLM} & \multirow{2}{*}{LLM} & \multirow{2}{*}{Architecture} & \multirow{2}{*}{Size} \\
      \cmidrule(lr){2-4}
       & Client 1      & Client 2      & Client 3                 &                         &             &       \\
      \midrule
      Scenario 1         & Gemma-2-2B    & Gemma-2-27B   & LLaDa-3-8B               & Qwen3-32B               & Different   & Different \\
      Scenario 2         & Gemma-2-2B    & LLaDa-3-8B    & Qwen3-32B                & Qwen2.5-72B-Instruct    & Different   & Different \\
      Scenario 3         & Qwen3-0.6B    & Qwen3-1.7B    & Qwen3-4B                 & Qwen3-32B               & Same        & Different \\
      Scenario 4         & Qwen3-8B      & LLaDa-3-8B    & Internlm3-8B-Instruct    & Qwen3-32B               & Different   & Same \\
      \bottomrule
    \end{tabular}
\end{table*}
\section{Experiments}
\subsection{Experimental Setup}
\label{Experimental Setup}
\noindent \textbf{Datasets:} We evaluate our proposed \textit{LaDa} on two well-established reasoning datasets: MATHInstruct \cite{yue2023mammoth} and CoT-Collection \cite{kim-etal-2023-cot}. MathInstruct is compiled from 13 math rationale datasets, containing problem-solution pairs with step-by-step reasoning chains across various mathematical domains including arithmetic, algebra, geometry, and calculus. CoT-Collection aggregates chain-of-thought reasoning instances from multiple sources including GSM8K, MATH, and TheoremQA, covering mathematical problem-solving, commonsense reasoning, and logical inference. While both datasets contain diverse task formats including open-ended questions, equation solving, and proof generation, we focus specifically on multiple-choice questions in our experiments to evaluate reasoning capacity transfer in a controlled setting.

\noindent \textbf{Baselines:} We compare \textit{LaDa} against four baseline methods:
\begin{itemize}
    \item \textbf{Standalone:} it trains each client-side SLM independently using only local private data without any knowledge transfer from server-side LLM, serving as a lower-bound baseline without collaborative learning.
    \item \textbf{FedKD \cite{seo2020federated}:} it represents classical federated knowledge distillation, where each client first trains its SLM on local data, then fine-tunes it using the LLM's chain-of-thought responses on the complete public dataset.
    \item \textbf{FedMKT \cite{fan2025fedmkt}:} it enables mutual knowledge transfer between server-side LLM and client-side SLMs through a public dataset. It employs a selective knowledge transfer mechanism based on minimum edit distance token alignment and dual minimum cross-entropy selection, filtering knowledge before model fine-tuning to ensure only beneficial knowledge is transferred bidirectionally.
    \item \textbf{WASP \cite{zoucontrastive}:} it generates synthetic datasets from pre-trained large language model and uses a voting mechanism to identify high-quality and low-quality samples based on their similarity to local private data.
\end{itemize}

\noindent \textbf{Scenarios:} We evaluate \textit{LaDa} across four federated large-small model collaboration scenarios with varying model heterogeneity, as detailed in Tab. \ref{tab:1}. Scenario 1 deploys a Qwen3-32B LLM with three heterogeneous SLMs: Gemma-2-2B, Gemma-2-27B, and LLaDa-3-8B, where both model architectures and sizes differ across clients. Scenario 2 employs a larger Qwen2.5-72B-Instruct LLM with Gemma-2-2B, LLaDa-3-8B, and Qwen3-32B SLMs to assess \textit{LaDa}'s performance when LLM has significantly more parameters. Scenario 3 focuses on architectural heterogeneity while maintaining size homogeneity, deploying three 8B-parameter SLMs with different model families: Qwen3-8B, LLaDa-3-8B, and InternLM3-8B-Instruct. Scenario 4 examines size heterogeneity under architectural homogeneity, utilizing Qwen3 SLMs with varying parameter scales of 0.6B\footnote{\hyperlink{https://huggingface.co/Qwen/Qwen3-0.6B}{https://huggingface.co/Qwen/Qwen3-0.6B}}, 1.7B\footnote{\hyperlink{https://huggingface.co/Qwen/Qwen3-1.7B}{https://huggingface.co/Qwen/Qwen3-1.7B}}, and 4B\footnote{\hyperlink{https://huggingface.co/Qwen/Qwen3-4B}{https://huggingface.co/Qwen/Qwen3-4B}} alongside a Qwen3-32B LLM. These scenarios comprehensively evaluate \textit{LaDa}'s adaptability across diverse collaborations.

\noindent \textbf{Implementation Details:} We extract 5,500 multiple-choice questions from the datasets, ensuring balanced distribution across all four answer options. Each of three clients holds 1,000 local training samples without chain-of-thought annotations and 500 local test samples. A public dataset containing 1,000 questions without answers is shared between the server and clients for knowledge transfer. The server's performance is evaluated on the combined test sets from all clients, totaling 1,500 questions. In the federated learning process, we conduct up to 5 communication rounds. Within each round, the training epochs range from 1 to 5 depending on the specific model configuration, with learning rates varying between 2e-4 and 9e-7. The MDP loss coefficient $\beta$ is set to 0.1. For the data filter module, we use a learning rate of 3e-4 and train for 10 epochs. All models are implemented using the PyTorch\footnote{\hyperlink{https://pytorch.org}{https://pytorch.org}} framework with DeepSpeed\footnote{\hyperlink{https://www.deepspeed.ai}{https://www.deepspeed.ai}} library for training acceleration, and experiments are conducted on 8 NVIDIA A800 GPUs.

\noindent \textbf{Evaluation:} We evaluate all models under both one-shot and zero-shot settings to assess their reasoning capabilities \cite{kojima2022large,kong2024better}. The distinction lies in whether the prompt includes a demonstration example: one-shot setting provides one standard CoT example in the prompt, while zero-shot setting contains no examples. Model performance is measured by accuracy on the test sets. Communication cost is quantified in MB to evaluate the efficiency of knowledge transfer between the server and clients. For the main experiments, we run each configuration three times and report the average performance.

\begin{table*}[t]
  \centering
  \caption{Main results on the MathInstruct dataset under one-shot and zero-shot settings.}
  \label{tab:2}
  \resizebox{\textwidth}{!}{
  \begin{tabular}{llcccccccccccc}
    \toprule
    \multicolumn{2}{c}{\multirow{2}{*}{Scenario}} & \multicolumn{6}{c}{One-shot setting} & \multicolumn{6}{c}{Zero-shot setting} \\
    \cmidrule(lr){3-8} \cmidrule(lr){9-14}
    & & Standalone & FedKD & FedMKT & WASP & \cellcolor{gray!30}Ours & Imp.(\%) & Standalone & FedKD & FedMKT & WASP & \cellcolor{gray!30}Ours & Imp.(\%) \\
    \midrule
    \multirow{4}{*}{Scenario 1}
      & Client 1 & 22.0$\scriptscriptstyle (\pm0.2)$ & 19.8$\scriptscriptstyle (\pm0.5)$ & \underline{22.8$\scriptscriptstyle (\pm0.1)$} & 21.4$\scriptscriptstyle (\pm0.4)$ & \cellcolor{gray!30}25.8$\scriptscriptstyle (\pm0.5)$ & 3.0 & 17.8$\scriptscriptstyle (\pm0.3)$ & 24.0$\scriptscriptstyle (\pm0.8)$ & 26.0$\scriptscriptstyle (\pm0.6)$ & \underline{26.4$\scriptscriptstyle (\pm0.5)$} & \cellcolor{gray!30}27.2$\scriptscriptstyle (\pm0.4)$ & 0.8 \\
      & Client 2 & 41.0$\scriptscriptstyle (\pm0.9)$ & 30.8$\scriptscriptstyle (\pm0.7)$ & 58.2$\scriptscriptstyle (\pm1.0)$ & \underline{58.8$\scriptscriptstyle (\pm0.6)$} & \cellcolor{gray!30}68.4$\scriptscriptstyle (\pm0.9)$ & 9.6 & 41.2$\scriptscriptstyle (\pm0.8)$ & 46.0$\scriptscriptstyle (\pm0.7)$ & \underline{46.2$\scriptscriptstyle (\pm0.6)$} & 44.0$\scriptscriptstyle (\pm0.5)$ & \cellcolor{gray!30}48.6$\scriptscriptstyle (\pm0.9)$ & 2.4 \\
      & Client 3 & 43.5$\scriptscriptstyle (\pm0.6)$ & 43.3$\scriptscriptstyle (\pm0.5)$ & 38.6$\scriptscriptstyle (\pm0.7)$ & \underline{51.2$\scriptscriptstyle (\pm1.1)$} & \cellcolor{gray!30}52.2$\scriptscriptstyle (\pm0.1)$ & 1.0 & 43.4$\scriptscriptstyle (\pm0.6)$ & 46.0$\scriptscriptstyle (\pm0.6)$ & 37.6$\scriptscriptstyle (\pm0.5)$ & \underline{46.4$\scriptscriptstyle (\pm0.7)$} & \cellcolor{gray!30}47.0$\scriptscriptstyle (\pm0.6)$ & 0.6 \\
      & LLM      & 42.7$\scriptscriptstyle (\pm0.4)$ & \underline{43.3$\scriptscriptstyle (\pm0.4)$} & 43.1$\scriptscriptstyle (\pm0.3)$ & \underline{43.3$\scriptscriptstyle (\pm0.3)$} & \cellcolor{gray!30}45.9$\scriptscriptstyle (\pm0.3)$ & 2.6 & 26.2$\scriptscriptstyle (\pm0.4)$ & 27.5$\scriptscriptstyle (\pm0.5)$ & 28.5$\scriptscriptstyle (\pm0.6)$ & \underline{29.4$\scriptscriptstyle (\pm0.5)$} & \cellcolor{gray!30}30.6$\scriptscriptstyle (\pm0.6)$ & 1.2 \\
    \midrule
    \multirow{4}{*}{Scenario 2}
      & Client 1 & 19.2$\scriptscriptstyle (\pm0.5)$ & 21.6$\scriptscriptstyle (\pm0.4)$ & 20.4$\scriptscriptstyle (\pm0.4)$ & \underline{21.8$\scriptscriptstyle (\pm0.5)$} & \cellcolor{gray!30}22.2$\scriptscriptstyle (\pm0.3)$ & 0.4 & 20.6$\scriptscriptstyle (\pm0.5)$ & 23.6$\scriptscriptstyle (\pm0.6)$ & 18.0$\scriptscriptstyle (\pm0.4)$ & \underline{23.8$\scriptscriptstyle (\pm0.7)$} & \cellcolor{gray!30}24.6$\scriptscriptstyle (\pm0.5)$ & 0.8 \\
      & Client 2 & 39.4$\scriptscriptstyle (\pm0.8)$ & 36.4$\scriptscriptstyle (\pm0.7)$ & 42.4$\scriptscriptstyle (\pm0.9)$ & \underline{44.2$\scriptscriptstyle (\pm1.0)$} & \cellcolor{gray!30}45.6$\scriptscriptstyle (\pm0.9)$ & 1.4 & 37.2$\scriptscriptstyle (\pm0.7)$ & 37.6$\scriptscriptstyle (\pm0.6)$ & 37.8$\scriptscriptstyle (\pm0.6)$ & \underline{38.0$\scriptscriptstyle (\pm0.5)$} & \cellcolor{gray!30}38.2$\scriptscriptstyle (\pm0.6)$ & 0.2 \\
      & Client 3 & 45.0$\scriptscriptstyle (\pm0.7)$ & 46.0$\scriptscriptstyle (\pm0.7)$ & \underline{46.8$\scriptscriptstyle (\pm0.8)$} & 46.6$\scriptscriptstyle (\pm0.7)$ & \cellcolor{gray!30}48.4$\scriptscriptstyle (\pm0.9)$ & 1.6 & 26.8$\scriptscriptstyle (\pm0.5)$ & 28.6$\scriptscriptstyle (\pm0.6)$ & 28.8$\scriptscriptstyle (\pm0.5)$ & \underline{29.0$\scriptscriptstyle (\pm0.6)$} & \cellcolor{gray!30}32.8$\scriptscriptstyle (\pm0.8)$ & 3.8 \\
      & LLM      & 83.1$\scriptscriptstyle (\pm0.6)$ & 83.3$\scriptscriptstyle (\pm0.7)$ & \underline{83.4$\scriptscriptstyle (\pm0.5)$} & 82.8$\scriptscriptstyle (\pm0.6)$ & \cellcolor{gray!30}83.8$\scriptscriptstyle (\pm0.5)$ & 0.4 & \underline{50.5$\scriptscriptstyle (\pm0.9)$} & 49.2$\scriptscriptstyle (\pm0.8)$ & 49.2$\scriptscriptstyle (\pm0.7)$ & 49.5$\scriptscriptstyle (\pm0.8)$ & \cellcolor{gray!30}54.2$\scriptscriptstyle (\pm1.0)$ & 4.7 \\
    \midrule
    \multirow{4}{*}{Scenario 3}
      & Client 1 & 34.4$\scriptscriptstyle (\pm0.6)$ & \underline{43.4$\scriptscriptstyle (\pm0.9)$} & 37.8$\scriptscriptstyle (\pm0.7)$ & 38.4$\scriptscriptstyle (\pm0.6)$ & \cellcolor{gray!30}44.8$\scriptscriptstyle (\pm1.0)$ & 1.4 & 31.6$\scriptscriptstyle (\pm0.6)$ & 30.2$\scriptscriptstyle (\pm0.5)$ & \underline{32.8$\scriptscriptstyle (\pm0.7)$} & 31.2$\scriptscriptstyle (\pm0.5)$ & \cellcolor{gray!30}35.0$\scriptscriptstyle (\pm0.8)$ & 2.2 \\
      & Client 2 & 67.6$\scriptscriptstyle (\pm1.1)$ & 68.6$\scriptscriptstyle (\pm1.0)$ & 68.6$\scriptscriptstyle (\pm0.9)$ & \underline{69.6$\scriptscriptstyle (\pm0.8)$} & \cellcolor{gray!30}70.4$\scriptscriptstyle (\pm0.9)$ & 0.8 & 40.8$\scriptscriptstyle (\pm0.7)$ & \underline{42.0$\scriptscriptstyle (\pm0.6)$} & 41.0$\scriptscriptstyle (\pm0.6)$ & 40.6$\scriptscriptstyle (\pm0.5)$ & \cellcolor{gray!30}43.0$\scriptscriptstyle (\pm0.7)$ & 1.0 \\
      & Client 3 & 80.6$\scriptscriptstyle (\pm1.3)$ & 81.6$\scriptscriptstyle (\pm1.2)$ & \underline{82.0$\scriptscriptstyle (\pm1.1)$} & 81.8$\scriptscriptstyle (\pm1.1)$ & \cellcolor{gray!30}83.2$\scriptscriptstyle (\pm1.2)$ & 1.2 & 47.6$\scriptscriptstyle (\pm0.9)$ & 48.4$\scriptscriptstyle (\pm0.9)$ & 49.0$\scriptscriptstyle (\pm0.8)$ & \underline{50.6$\scriptscriptstyle (\pm1.0)$} & \cellcolor{gray!30}64.4$\scriptscriptstyle (\pm1.6)$ & 13.8 \\
      & LLM      & 42.1$\scriptscriptstyle (\pm0.5)$ & 43.9$\scriptscriptstyle (\pm0.6)$ & 44.1$\scriptscriptstyle (\pm0.5)$ & \underline{46.0$\scriptscriptstyle (\pm0.7)$} & \cellcolor{gray!30}49.8$\scriptscriptstyle (\pm0.8)$ & 3.8 & 25.8$\scriptscriptstyle (\pm0.4)$ & 26.1$\scriptscriptstyle (\pm0.4)$ & 26.3$\scriptscriptstyle (\pm0.5)$ & \underline{26.7$\scriptscriptstyle (\pm0.5)$} & \cellcolor{gray!30}27.8$\scriptscriptstyle (\pm0.5)$ & 1.1 \\
    \midrule
    \multirow{4}{*}{Scenario 4}
      & Client 1 & 79.6$\scriptscriptstyle (\pm0.9)$ & 80.0$\scriptscriptstyle (\pm0.8)$ & \underline{80.6$\scriptscriptstyle (\pm0.9)$} & \underline{80.6$\scriptscriptstyle (\pm0.9)$} & \cellcolor{gray!30}82.6$\scriptscriptstyle (\pm1.0)$ & 2.0 & 61.0$\scriptscriptstyle (\pm1.0)$ & 68.8$\scriptscriptstyle (\pm1.1)$ & 69.6$\scriptscriptstyle (\pm1.0)$ & \underline{71.0$\scriptscriptstyle (\pm1.0)$} & \cellcolor{gray!30}72.4$\scriptscriptstyle (\pm1.1)$ & 1.4 \\
      & Client 2 & 40.8$\scriptscriptstyle (\pm0.8)$ & 43.8$\scriptscriptstyle (\pm0.9)$ & 41.0$\scriptscriptstyle (\pm0.7)$ & \underline{47.0$\scriptscriptstyle (\pm1.0)$} & \cellcolor{gray!30}48.0$\scriptscriptstyle (\pm1.1)$ & 1.0 & 40.2$\scriptscriptstyle (\pm0.8)$ & \underline{45.4$\scriptscriptstyle (\pm0.9)$} & 44.0$\scriptscriptstyle (\pm0.8)$ & 43.2$\scriptscriptstyle (\pm0.7)$ & \cellcolor{gray!30}46.2$\scriptscriptstyle (\pm0.9)$ & 0.8 \\
      & Client 3 & 51.6$\scriptscriptstyle (\pm0.8)$ & \underline{53.8$\scriptscriptstyle (\pm0.9)$} & 52.2$\scriptscriptstyle (\pm0.8)$ & 52.8$\scriptscriptstyle (\pm0.8)$ & \cellcolor{gray!30}55.0$\scriptscriptstyle (\pm0.9)$ & 1.2 & 56.8$\scriptscriptstyle (\pm0.9)$ & 52.2$\scriptscriptstyle (\pm0.7)$ & \underline{57.8$\scriptscriptstyle (\pm1.0)$} & 57.4$\scriptscriptstyle (\pm1.0)$ & \cellcolor{gray!30}59.0$\scriptscriptstyle (\pm1.1)$ & 1.2 \\
      & LLM      & 42.7$\scriptscriptstyle (\pm0.5)$ & 35.9$\scriptscriptstyle (\pm0.6)$ & 45.7$\scriptscriptstyle (\pm0.7)$ & 43.4$\scriptscriptstyle (\pm0.5)$ & \cellcolor{gray!30}47.4$\scriptscriptstyle (\pm0.7)$ & 1.7 & 25.6$\scriptscriptstyle (\pm0.4)$ & 25.8$\scriptscriptstyle (\pm0.4)$ & \underline{26.0$\scriptscriptstyle (\pm0.5)$} & \underline{26.0$\scriptscriptstyle (\pm0.5)$} & \cellcolor{gray!30}28.5$\scriptscriptstyle (\pm0.6)$ & 2.5 \\
    \bottomrule
  \end{tabular}}
\end{table*}

\begin{table*}[t]
  \centering
  \caption{Main results on the CoT-Collection dataset under one-shot and zero-shot settings.}
  \label{tab:3}
  \resizebox{\textwidth}{!}{
  \begin{tabular}{llcccccccccccc}
    \toprule
    \multicolumn{2}{c}{\multirow{2}{*}{Scenario}} & \multicolumn{6}{c}{One-shot setting} & \multicolumn{6}{c}{Zero-shot setting} \\
    \cmidrule(lr){3-8} \cmidrule(lr){9-14}
    & & Standalone & FedKD & FedMKT & WASP & \cellcolor{gray!30}Ours & Imp.(\%) & Standalone & FedKD & FedMKT & WASP & \cellcolor{gray!30}Ours & Imp.(\%) \\
    \midrule
    \multirow{4}{*}{Scenario 1}
      & Client 1 & 40.8$\scriptscriptstyle (\pm0.6)$ & 41.6$\scriptscriptstyle (\pm0.3)$ & 42.8$\scriptscriptstyle (\pm0.8)$ & \underline{44.0$\scriptscriptstyle (\pm0.5)$} & \cellcolor{gray!30}47.8$\scriptscriptstyle (\pm0.9)$ & 3.8 & 44.6$\scriptscriptstyle (\pm0.6)$ & 44.6$\scriptscriptstyle (\pm0.4)$ & 45.2$\scriptscriptstyle (\pm0.7)$ & \underline{47.6$\scriptscriptstyle (\pm0.8)$} & \cellcolor{gray!30}49.6$\scriptscriptstyle (\pm1.0)$ & 2.0 \\
      & Client 2 & 64.9$\scriptscriptstyle (\pm1.0)$ & 66.2$\scriptscriptstyle (\pm0.7)$ & 69.2$\scriptscriptstyle (\pm1.2)$ & \underline{69.8$\scriptscriptstyle (\pm0.4)$} & \cellcolor{gray!30}70.0$\scriptscriptstyle (\pm0.8)$ & 0.2 & 47.0$\scriptscriptstyle (\pm0.9)$ & 48.6$\scriptscriptstyle (\pm0.5)$ & 48.6$\scriptscriptstyle (\pm0.6)$ & \underline{65.6$\scriptscriptstyle (\pm1.3)$} & \cellcolor{gray!30}69.0$\scriptscriptstyle (\pm1.4)$ & 3.4 \\
      & Client 3 & 41.0$\scriptscriptstyle (\pm0.4)$ & 43.8$\scriptscriptstyle (\pm0.9)$ & 41.6$\scriptscriptstyle (\pm0.5)$ & \underline{50.2$\scriptscriptstyle (\pm1.1)$} & \cellcolor{gray!30}51.2$\scriptscriptstyle (\pm0.7)$ & 1.0 & 43.2$\scriptscriptstyle (\pm0.8)$ & 44.2$\scriptscriptstyle (\pm0.6)$ & 43.6$\scriptscriptstyle (\pm0.5)$ & \underline{52.0$\scriptscriptstyle (\pm1.0)$} & \cellcolor{gray!30}54.0$\scriptscriptstyle (\pm1.2)$ & 2.0 \\
      & LLM      & 50.6$\scriptscriptstyle (\pm0.7)$ & 51.3$\scriptscriptstyle (\pm0.6)$ & 52.5$\scriptscriptstyle (\pm0.8)$ & \underline{52.9$\scriptscriptstyle (\pm0.5)$} & \cellcolor{gray!30}54.0$\scriptscriptstyle (\pm0.9)$ & 1.1 & 39.6$\scriptscriptstyle (\pm0.5)$ & 40.1$\scriptscriptstyle (\pm0.3)$ & 41.9$\scriptscriptstyle (\pm0.6)$ & \underline{42.7$\scriptscriptstyle (\pm0.7)$} & \cellcolor{gray!30}43.5$\scriptscriptstyle (\pm0.8)$ & 0.8 \\
    \bottomrule
  \end{tabular}}
\end{table*}

\subsection{Overall Performance}
Tab. \ref{tab:2} and Tab. \ref{tab:3} present the main experimental results on the MathInstruct and CoT-Collection datasets, respectively. The underlined values indicate the best performance among baseline methods for each configuration. Our proposed \textit{LaDa} consistently outperforms all baseline methods across all four scenarios under both one-shot and zero-shot settings. The improvements over the best-performing baseline range from 0.2\% to 13.8\%, demonstrating the effectiveness of our bidirectional reasoning ability distillation approach. Specifically, on the MathInstruct dataset under scenario 1 with one-shot setting, client 2 achieves a 9.6\% improvement while the server LLM gains 2.6\%, demonstrating mutual benefits from bidirectional knowledge transfer. Scenario 3 under zero-shot setting shows even more substantial gains, with client 3 improving by 13.8\%. This indicates that \textit{LaDa} is particularly effective when transferring reasoning capacity among models within the same architecture family but with different parameter scales. On the CoT-Collection dataset, \textit{LaDa} maintains consistent improvements across all configurations. In scenario 1, client 1 achieves a 3.8\% gain under one-shot setting while client 2 demonstrates a 3.4\% improvement under zero-shot setting. These results validate that \textit{LaDa} generalizes well across different datasets. Comparing the one-shot and zero-shot settings, we find that \textit{LaDa} delivers substantial gains in both configurations. While models generally perform better with demonstration examples in the one-shot setting, the zero-shot results highlight \textit{LaDa}'s ability to enhance the intrinsic reasoning capabilities of both client-side SLMs and the server-side LLM without relying on in-context examples. Finally, an important observation across all scenarios is that LLM consistently benefits from knowledge transfer from SLMs, not just the reverse. This bidirectional improvement validates our design choice of enabling mutual knowledge exchange rather than unidirectional distillation from the server to clients.

\subsection{Reasoning Capacity Comparison}
To further validate the effectiveness of our model adaptive reasoning distillation approach in transferring reasoning capabilities, we conduct experiments against three representative reasoning enhancement methods including SFT \cite{chung2024scaling}, SOCRATIC COT \cite{shridhar2023distilling}, and rStar \cite{qi2024mutual} under scenario 1 zero-shot setting, as presented in Tab. \ref{tab:reasoning}. The results demonstrate that LaDa achieves substantial improvements over all baseline methods across client models. Specifically, LaDa outperforms the best baseline by 3.2\% for Client 1, achieving 25.8\% accuracy compared to 22.6\% from rStar. For Client 2, LaDa delivers the most significant gain with 68.4\% accuracy. Client 3 shows a modest improvement, with LaDa reaching 52.2\% accuracy compared to 51.6\% from rStar. The consistent improvements across all three clients validate that our model adaptive reasoning distillation approach effectively transfers reasoning patterns from the server LLM to heterogeneous client SLMs, enabling them to learn step-by-step reasoning abilities within their individual capacity constraints.

\begin{table}[t]
\centering
\caption{Reasoning capacity comparison results.}
\label{tab:reasoning}
\begin{tabular}{lcccc}
\toprule
\multirow{2}{*}{Model} & \multicolumn{4}{c}{Scenario 1 zero-shot setting} \\
\cmidrule(lr){2-5}
& Client 1 & Client 2 & Client 3 & LLM \\
\midrule
SFT & 22.0 & 41.8 & 43.2 & - \\
SOCRATIC COT & 21.4 & 45.4 & 47.0 & - \\
rStar & 22.6 & 54.4 & 51.6 & - \\
Ours & 25.8 & 68.4 & 52.2 & 45.9 \\
\bottomrule
\end{tabular}
\end{table}

\begin{table*}[t]
  \centering
  \caption{Ablation study results on the MathInstruct dataset.}
  \label{tab:ablation}
  \resizebox{\textwidth}{!}{
  \begin{tabular}{llcccccccc}
    \toprule
    \multirow{2}{*}{Setting} & \multirow{2}{*}{Method} & \multicolumn{4}{c}{Scenario 1} & \multicolumn{4}{c}{Scenario 2} \\
    \cmidrule(lr){3-6} \cmidrule(lr){7-10}
    & & Client 1 & Client 2 & Client 3 & LLM & Client 1 & Client 2 & Client 3 & LLM \\
    \midrule
    \multirow{4}{*}{One-shot setting}
      & \cellcolor{gray!30}Ours     & \cellcolor{gray!30}26.4 & \cellcolor{gray!30}61.4 & \cellcolor{gray!30}58.6 & \cellcolor{gray!30}44.4 & \cellcolor{gray!30}22.2 & \cellcolor{gray!30}45.6 & \cellcolor{gray!30}48.4 & \cellcolor{gray!30}83.8 \\
      & \quad w/o reasoning distillation  & 21.8$\scriptscriptstyle (-4.6)$ & 53.4$\scriptscriptstyle (-8.0)$ & 44.8$\scriptscriptstyle (-13.8)$ & 44.0$\scriptscriptstyle (-0.4)$ & 21.0$\scriptscriptstyle (-1.2)$ & 40.0$\scriptscriptstyle (-5.6)$ & 47.4$\scriptscriptstyle (-1.0)$ & 83.1$\scriptscriptstyle (-0.7)$ \\
      & \quad w/o filter   & 22.8$\scriptscriptstyle (-3.6)$ & 58.2$\scriptscriptstyle (-3.2)$ & 38.6$\scriptscriptstyle (-20.0)$ & 42.8$\scriptscriptstyle (-1.6)$ & 20.4$\scriptscriptstyle (-1.8)$ & 42.4$\scriptscriptstyle (-3.2)$ & 46.8$\scriptscriptstyle (-1.6)$ & 83.4$\scriptscriptstyle (-0.4)$ \\
      & \quad w/o sample weights   & 25.4$\scriptscriptstyle (-1.0)$ & 58.7$\scriptscriptstyle (-2.7)$ & 52.6$\scriptscriptstyle (-6.0)$ & 43.1$\scriptscriptstyle (-1.3)$ & 21.6$\scriptscriptstyle (-0.6)$ & 41.0$\scriptscriptstyle (-4.6)$ & 27.6$\scriptscriptstyle (-20.8)$ & 83.0$\scriptscriptstyle (-0.8)$ \\
    \midrule
    \multirow{4}{*}{Zero-shot setting}
      & \cellcolor{gray!30}Ours  & \cellcolor{gray!30}27.2 & \cellcolor{gray!30}48.6 & \cellcolor{gray!30}47.0 & \cellcolor{gray!30}30.6 & \cellcolor{gray!30}24.6 & \cellcolor{gray!30}38.2 & \cellcolor{gray!30}32.8 & \cellcolor{gray!30}54.2 \\
      & \quad w/o reasoning distillation  & 20.6$\scriptscriptstyle (-6.6)$ & 38.2$\scriptscriptstyle (-10.4)$ & 39.2$\scriptscriptstyle (-7.8)$ & 26.3$\scriptscriptstyle (-4.3)$ & 21.2$\scriptscriptstyle (-3.4)$ & 38.2$\scriptscriptstyle (0.0)$ & 29.8$\scriptscriptstyle (-3.0)$ & 54.0$\scriptscriptstyle (-0.2)$ \\
      & \quad w/o filter   & 26.0$\scriptscriptstyle (-1.2)$ & 46.2$\scriptscriptstyle (-2.4)$ & 37.6$\scriptscriptstyle (-9.4)$ & 28.5$\scriptscriptstyle (-2.1)$ & 18.0$\scriptscriptstyle (-6.6)$ & 37.8$\scriptscriptstyle (-0.4)$ & 28.8$\scriptscriptstyle (-4.0)$ & 49.2$\scriptscriptstyle (-5.0)$ \\
      & \quad w/o sample weights   & 25.7$\scriptscriptstyle (-1.5)$ & 46.8$\scriptscriptstyle (-1.8)$ & 46.2$\scriptscriptstyle (-0.8)$ & 29.0$\scriptscriptstyle (-1.6)$ & 21.0$\scriptscriptstyle (-3.6)$ & 36.4$\scriptscriptstyle (-1.8)$ & 26.6$\scriptscriptstyle (-6.2)$ & 53.6$\scriptscriptstyle (-0.6)$ \\
    \bottomrule
  \end{tabular}}
\end{table*}
\begin{figure*}[t] 
\centering
\includegraphics[width=1.0\linewidth]{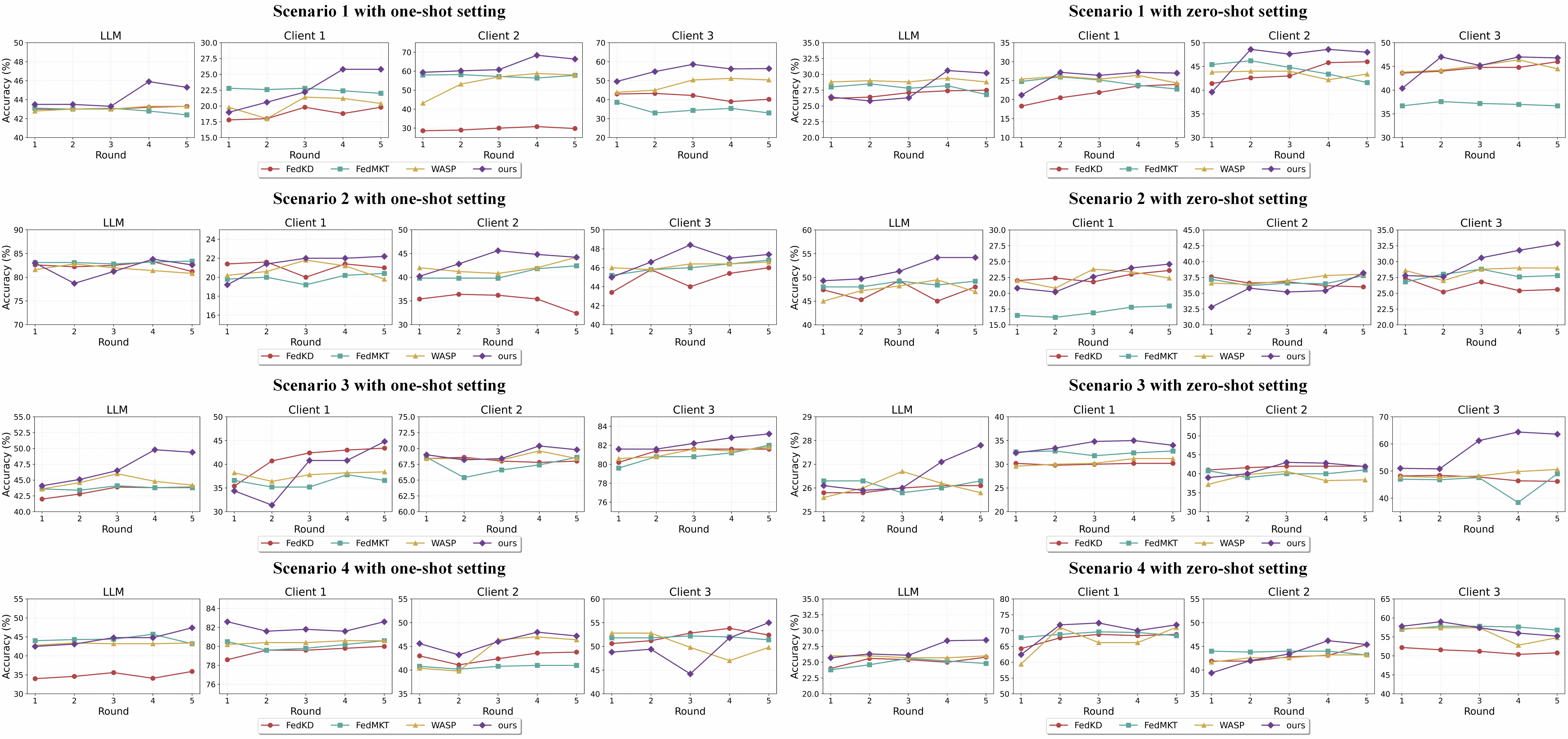}
\caption{Convergence analysis results of our proposed ReDA.}
\label{fig:convergence}
\end{figure*}
\subsection{Ablation Study}
To validate the effectiveness of each component in \textit{LaDa}, we conduct ablation study on the MathInstruct dataset across scenario 1 and scenario 2 under both one-shot and zero-shot settings. We evaluate three ablation variants: 1) w/o reasoning distillation, where we replace the model adaptive reasoning distillation loss $\mathcal{L}^{\text{mard}}_{\text{client}}$ and $\mathcal{L}^{\text{mard}}_{\text{server}}$ with standard cross-entropy loss; 2) w/o filter, where we remove model learnability-aware sample filter and directly use all samples from the public dataset without any filtering; 3) w/o overlapping sample weights, where we eliminate the sample weighting mechanism on the server side and treat all samples equally during loss computation. Tab. \ref{tab:ablation} presents the results. Specifically, removing the reasoning distillation component leads to substantial performance degradation. For example, in scenario 1 under one-shot setting, Client 2 and Client 3 drop by 8.0\% and 13.8\% respectively, validating that our reasoning loss is essential for transferring reasoning patterns between heterogeneous models. The absence of the sample filter also causes performance drops, with Client 3 in scenario 1 decreasing from 47.0\% to 37.6\% under zero-shot setting, indicating that selective filtering prevents low-quality samples from degrading model performance. Removing the sample weighting mechanism particularly affects certain scenarios, as shown by Client 3 in Scenario 2 dropping from 48.4\% to 27.6\%, demonstrating that proper weighting is necessary for balanced knowledge aggregation. Overall, these ablation study results confirm that each component of \textit{LaDa} contributes meaningfully to the overall performance.

\begin{figure*}[t] 
\centering
\includegraphics[width=0.9\linewidth]{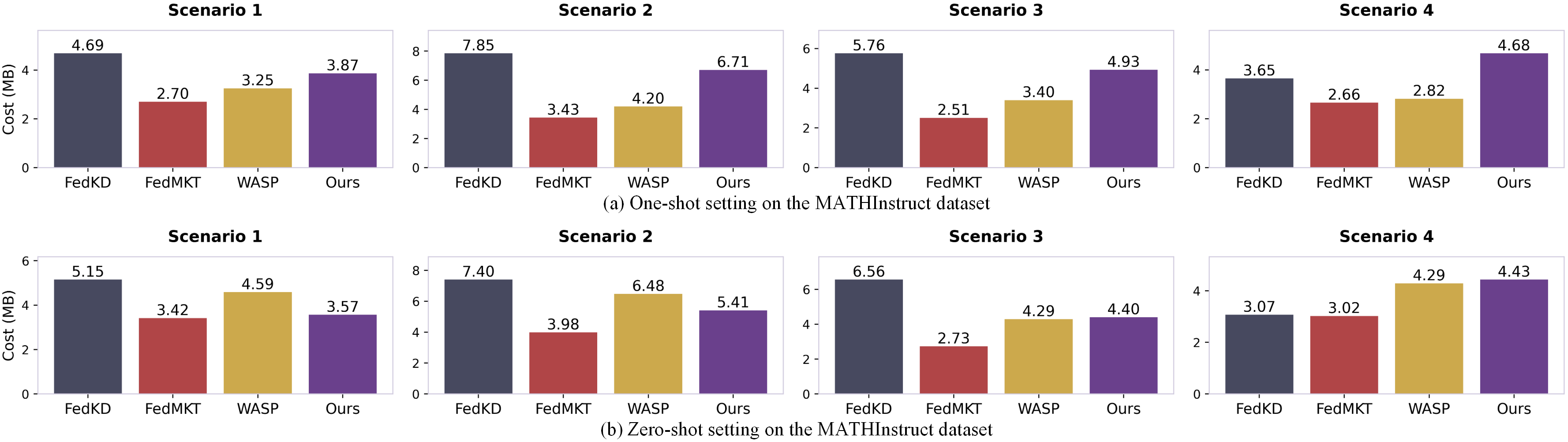}
\caption{Communication cost comparison of our proposed ReDA.}
\label{fig:cost}
\end{figure*}
\begin{table*}[t]
  \centering
  \caption{Hyperparameter analysis on scenario 1 under one-shot and zero-shot settings.}
  \label{tab:hyper}
  \begin{tabular}{lccccccc ccccc}
    \toprule
    \multirow{2}{*}{Dataset} & \multicolumn{2}{c}{Hyperparameter} & \multicolumn{5}{c}{Scenario 1 One-shot setting} & \multicolumn{5}{c}{Scenario 1 Zero-shot setting} \\
    \cmidrule(lr){2-3}\cmidrule(lr){4-8}\cmidrule(lr){9-13}
    & $\alpha$ & $\lambda$ & Client 1 & Client 2 & Client 3 & LLM & avg & Client 1 & Client 2 & Client 3 & LLM & avg \\
    \midrule
    \multirow{4}{*}{MathInstruct}
      & 0.33 & 0.33 & 24.4 & 61.4 & 58.6 & 44.4 & 47.2 & \cellcolor{gray!30}27.2 & \cellcolor{gray!30}48.6 & \cellcolor{gray!30}47.0 & \cellcolor{gray!30}30.6 & \cellcolor{gray!30}38.4 \\
      & 0.33 & 0.66 & 20.2 & 59.2 & 54.8 & 44.3 & 44.6 & 26.2 & 48.2 & 47.0 & 28.6 & 37.5 \\
      & 0.66 & 0.33 & 20.0 & 60.8 & 55.6 & 43.0 & 44.9 & 25.6 & 48.6 & 43.0 & 26.6 & 36.0 \\
      & 0.66 & 0.66 & \cellcolor{gray!30}25.8 & \cellcolor{gray!30}68.4 & \cellcolor{gray!30}52.2 & \cellcolor{gray!30}45.9 & \cellcolor{gray!30}48.1 & 26.4 & 49.0 & 38.7 & 29.2 & 35.8 \\
    \midrule
    \multirow{4}{*}{CoT-Collection}
      & 0.33 & 0.33 & 46.0 & 69.4 & 51.0 & 52.1 & 54.6 & \cellcolor{gray!30}59.6 & \cellcolor{gray!30}69.0 & \cellcolor{gray!30}54.0 & \cellcolor{gray!30}43.5 & \cellcolor{gray!30}56.5 \\
      & 0.33 & 0.66 & 48.0 & 69.4 & 50.0 & 51.6 & 54.8 & 51.4 & 71.0 & 53.2 & 42.5 & 54.5 \\
      & 0.66 & 0.33 & \cellcolor{gray!30}47.8 & \cellcolor{gray!30}70.0 & \cellcolor{gray!30}51.2 & \cellcolor{gray!30}54.0 & \cellcolor{gray!30}55.8 & 53.4 & 69.0 & 49.6 & 42.1 & 53.5 \\
      & 0.66 & 0.66 & 45.8 & 68.4 & 50.0 & 51.8 & 54.0 & 51.6 & 65.2 & 52.6 & 42.0 & 52.9 \\
    \bottomrule
  \end{tabular}
\end{table*}
\subsection{Convergence Experiments}
To evaluate the convergence behavior of \textit{LaDa}, we conduct experiments across all four scenarios under both one-shot and zero-shot settings on the MathInstruct dataset. Fig. \ref{fig:convergence} presents the accuracy curves over 5 communication rounds for the LLM and three SLMs, where each row represents one scenario, with the left four subplots showing one-shot setting results and the right four subplots showing zero-shot setting results. The results demonstrate that \textit{LaDa} achieves stable convergence and higher final accuracy compared to baseline methods. For instance, In scenario 1 with one-shot setting, most clients show steady improvement throughout the rounds. Client 1 demonstrates consistent upward trends from 19.0\% at round 1 to 25.8\% at round 5, while client 2 progresses from 59.4\% to 66.4\%. Under zero-shot setting, convergence patterns remain stable, with client 2 reaching 48.6\% by round 2 and maintaining similar performance. In scenario 2, which involves greater model heterogeneity, \textit{LaDa} shows different convergence characteristics. The LLM exhibits some fluctuations in one-shot setting, dropping from 83.0\% at round 1 to 78.7\% at round 2 before recovering to 83.8\% at round 4. This temporary decrease suggests an adaptation period as the model incorporates diverse reasoning patterns from heterogeneous SLMs. Finally, scenarios 3 and 4 present similar convergence trends, with most models reaching stable performance within 3 or 4 rounds. 
Notably, in scenario 3 zero-shot setting, client 3 experiences a significant jump from 50.8\% at round 2 to 61.2\% at round 3, then stabilizes around 64.4\%, indicating that effective reasoning patterns were successfully transferred during that round. 
The convergence analysis confirms that \textit{LaDa} achieves stable training process across diverse collaboration scenarios.

\subsection{Communication Cost}
Fig. \ref{fig:cost} presents the communication cost comparison across different methods under various scenarios and settings. Our analysis reveals that \textit{LaDa} achieves a favorable balance between performance gains and communication overhead. In the one-shot setting on the MathInstruct dataset, \textit{LaDa} maintains moderate communication costs across scenarios, with costs ranging from 3.87 MB to 6.71 MB. While FedKD generally incurs higher communication costs due to frequent model updates, FedMKT achieves the lowest costs by limiting knowledge transfer scope. In the zero-shot setting, \textit{LaDa} demonstrates improved communication efficiency. For instance, in scenario 1, our method requires only 3.57 MB compared to FedKD's 5.15 MB and WASP's 4.59 MB, while still achieving the best performance. Similarly, in scenario 3, \textit{LaDa} uses 4.40 MB, substantially less than FedKD's 6.56 MB, demonstrating efficient knowledge transfer. On the contrary, the communication costs in scenario 2 and scenario 4 are relatively higher across all methods due to the increased model heterogeneity and complexity of knowledge transfer. Overall, our proposed method achieves superior performance improvements as shown in the main experiments while keeping communication costs competitive with existing approaches.

\begin{figure*}[t] 
\centering
\includegraphics[width=0.9\linewidth]{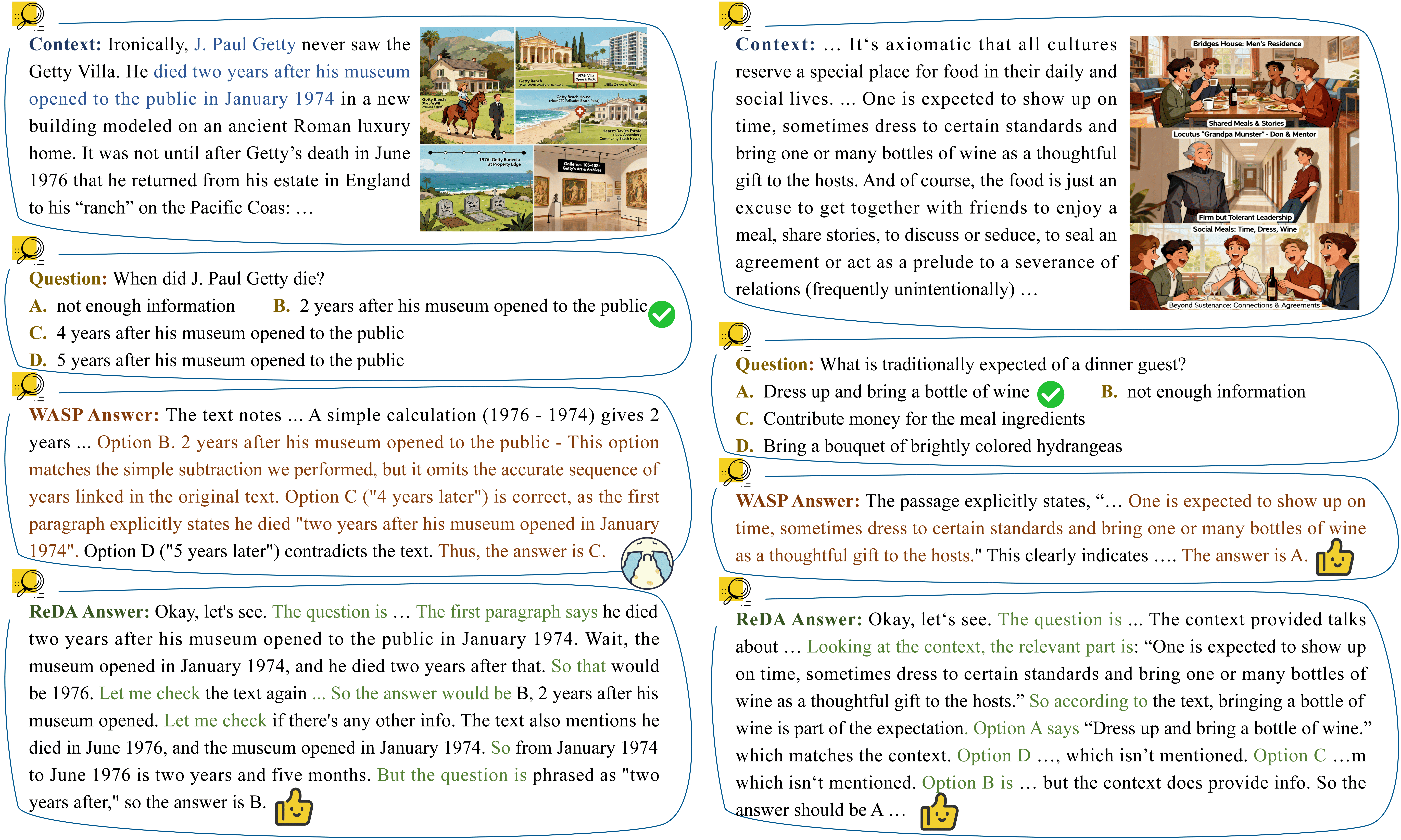}
\caption{Case study results.}
\label{fig:case}
\end{figure*}
\subsection{Hyperparameter Analysis}
We conduct hyperparameter analysis to study the impact of two key parameters in \textit{LaDa}: $\alpha$ and $\lambda$. The parameter $\alpha$ controls the weight of the batch-level and round-level rewards, while $\lambda$ balances the contribution of the reasoning distillation loss in the sample filter. Tab. \ref{tab:hyper} presents the results on Scenario 1 with different hyperparameter configurations. On the MathInstruct dataset, we observe that the optimal hyperparameter settings differ between one-shot and zero-shot configurations. Under the one-shot setting, $\alpha=0.66$ and $\lambda=0.66$ achieves the best average performance of 48.1\%, with particularly strong improvements on Client 2 achieving 68.4\% and the server LLM achieving 45.9\%. In contrast, the zero-shot setting favors $\alpha=0.33$ and $\lambda=0.33$, reaching an average performance of 38.4\% with more balanced improvements across all clients and the LLM. For the CoT-Collection dataset, the optimal configurations also vary across settings. The one-shot setting achieves the best average performance of 55.8\% with $\alpha=0.66$ and $\lambda=0.33$, while the zero-shot setting performs optimally with $\alpha=0.33$ and $\lambda=0.33$, reaching 56.5\%. Notably, both datasets consistently show that zero-shot settings benefit from smaller hyperparameter values, suggesting that more conservative knowledge transfer is preferable when models operate without in-context examples. Overall, these results demonstrate that \textit{LaDa} is relatively robust to hyperparameter choices, with performance remaining competitive across different configurations.

\subsection{Case Study}
To illustrate how \textit{LaDa} enables effective reasoning ability transfer between different large and small models, we present two representative cases from the MathInstruct dataset in Fig. \ref{fig:case}. The visual illustrations accompanying each context are generated using Seedream-4.0\footnote{\hyperlink{https://www.byteplus.com/en/product/Seedream}{https://www.byteplus.com/en/product/Seedream}} text-to-image model to better convey the contextual scenarios, as the original dataset contains only textual information without images. Specifically, the first case examines a question about when J. Paul Getty died, based on contextual information stating that the museum opened in January 1974 and Getty passed away two years thereafter. WASP demonstrates a fundamentally flawed reasoning process by initially calculating that 1976 minus 1974 equals 2 years, yet paradoxically concludes that option C stating four years is correct. The reasoning explicitly acknowledges that the text mentions Getty died two years after the museum opened, but then contradicts itself by selecting the four-year option. This logical inconsistency reveals that WASP fails to maintain coherence between its intermediate reasoning steps and final answer selection. In contrast, \textit{LaDa} exhibits a methodical reasoning chain by first identifying the museum opening date in January 1974, then calculating that two years later would be 1976, and subsequently validating this conclusion by cross-referencing with the explicit mention of Getty's death in June 1976. \textit{LaDa} correctly recognizes that the time span from January 1974 to June 1976 is approximately two years, leading to the selection of option B. This case demonstrates that \textit{LaDa}'s model adaptive reasoning distillation successfully transfers the ability to maintain logical consistency throughout the entire reasoning process, from premise identification through intermediate calculations to final conclusion. The second case addresses the question of what is traditionally expected of a dinner guest, with the context describing social dining customs including punctuality, appropriate attire, and bringing wine as a thoughtful gift. While WASP arrives at correct answer by directly quoting a relevant passage from the context, its reasoning process remains relatively superficial without logical analysis. However, \textit{LaDa} first parses the question requirements, then methodically examining the context to locate pertinent information about guest expectations. \textit{LaDa} explicitly evaluates each answer option against the textual evidence, systematically noting that option D regarding hydrangeas lacks textual support, option C about monetary contributions is not mentioned, and option B claiming insufficient information is invalid because the context clearly provides explicit expectations. These two cases demonstrate that \textit{LaDa} successfully transfers fundamental reasoning capabilities rather than surface-level answer patterns, enabling client models to tackle diverse question types with appropriate analytical rigor and logical coherence.

\section{Conclusion}
This paper propose a plug-in for existing large-small model collaboration frameworks. We identify two critical challenges in large-small model collaboration: (i) bidirectional model learnability gap, where SLMs and LLM cannot identify high-reward samples matching their learnability constraints; (ii) domain-agnostic reasoning transfer, where existing methods fail to flexible adapt to local domain to achieve reasoning transfer for LLM and SLM. Therefore, we propose a domain adaptive reasoning distillation method with model learnability-aware data filter, facilitating bidirectional knowledge transfer. We provide $O(1/\sqrt{T})$ convergence rate in a classic large-small model collaboration framework and demonstrate up to 13.8\% accuracy improvements over baselines through experiments across four collaboration scenarios with varying model architectures and scales on two widely-used datasets.

\section{AI-Generated Content Acknowledgement}
During the preparation of this work, the authors used DeepSeek-V3.2 to improve language and readability. Additionally, Seedream-4.0 was used to generate visual illustrations for the case study in Fig. \ref{fig:case}. After using these tools, the authors reviewed and edited the content as needed and take full responsibility for the content of the publication.

\bibliographystyle{IEEEtran}
\bibliography{software}

\end{document}